\documentclass[letterpaper,10pt,journal,twoside]{IEEEtran}
\usepackage{amsmath,amsfonts}
\usepackage{array}
\usepackage{textcomp}
\usepackage{stfloats}
\usepackage{url}
\usepackage{verbatim}
\usepackage{graphicx}
\usepackage{cite}
\usepackage{lipsum}
\usepackage{paralist}
\usepackage{makecell}

\usepackage{amsmath}
\usepackage{amssymb}
\usepackage{amsthm}
\usepackage{mathtools}
\usepackage{bm}
\usepackage{mathrsfs}
\usepackage[dvipsnames]{xcolor}
\usepackage{xfp}
\usepackage{pifont}

\usepackage[most]{tcolorbox}

\usepackage{siunitx}
\usepackage{tikz}
\usepackage{booktabs}
\usepackage{multirow}

\usetikzlibrary{arrows}
\usetikzlibrary{shapes}
\usetikzlibrary{snakes}
\usetikzlibrary{calc}
\usetikzlibrary{arrows.meta}
\usepackage{pgfplots}
\pgfplotsset{compat=1.18}
\usetikzlibrary{pgfplots.statistics, pgfplots.colorbrewer, pgfplots.groupplots}
\usetikzlibrary{positioning}
\usetikzlibrary{fit}
\usepackage{pgfplotstable}

\usepackage{ifthen}
\usepackage{etoolbox}
\usepackage{setspace}
\usepackage{soul}
\usepackage[hidelinks]{hyperref}

\usepackage{paralist}

\definecolor{lilac}{rgb}{0.78, 0.64, 0.78}

\makeatletter
\let\MYcaption\@makecaption
\makeatother

\usepackage[font=footnotesize]{subcaption}

\makeatletter
\let\@makecaption\MYcaption
\makeatother

\newcommand{\fakepar}[1]{\vspace{1mm}\noindent\textbf{#1}}

\newcommand{\R}{\mathbb{R}}
\newcommand{\K}{\mathcal{K}}
\newcommand{\C}{\mathcal{C}}

\DeclareMathOperator{\diag}{diag}
\DeclareMathOperator{\nnz}{nnz}

\newif\ifshowchanges
\showchangesfalse       %

\newcommand{\rev}[1]{%
  \begingroup
  \ifshowchanges\color{blue}\fi
  #1%
  \endgroup
}

\usepackage[ruled,vlined,noend,linesnumbered]{algorithm2e}

\usepackage{comment}

\newcommand{\toolboxname}{WarpMPC\xspace}

\newif\ifanonymized
\anonymizedfalse   %

\newif\ifarxiv
\arxivfalse

\AtBeginEnvironment{thebibliography}{%
\ifarxiv\setstretch{0.89}\else\setstretch{0.95}\fi
}

\AtBeginDocument{%
  \ifarxiv
    \AddToHookNext{shipout/foreground}{%
      \begin{tikzpicture}[remember picture,overlay]
        \node[anchor=south,draw=black,fill=white,inner sep=3pt,
              text width=\dimexpr\textwidth-6pt\relax,
              font=\fontsize{6}{7}\selectfont,align=left]
          at ([yshift=8pt]current page.south) {%
          \textcopyright\ 2026 IEEE. Personal use of this material is permitted. Permission from IEEE must be obtained for all other uses, in any current or future media, including reprinting/republishing this material for advertising or promotional purposes, creating new collective works, for resale or redistribution to servers or lists, or reuse of any copyrighted component of this work in other works.};
      \end{tikzpicture}%
    }%
  \fi
}

\anonymizedfalse\showchangesfalse\arxivtrue
\begin{document}
\title{\toolboxname: Large-Batch MPC on GPU \\via ADMM with Unrolled~$LDL^\top$ Factorization}
\ifanonymized
\author{Anonymous Authors%
\thanks{\vspace{2.5em}}%
\thanks{Anonymized.}%
\thanks{\vspace{2.5em}}
}
\else
\author{Henrik Hose$^{1,2}$, Se Hwan Jeon$^2$, Charles Khazoom$^2$, Sangbae Kim$^2$, Sebastian Trimpe$^1$%
    \thanks{Manuscript received: July 13, 2026; Revised: October 8, 2026; Accepted: November 9, 2026. This paper was recommended for publication by Editor Lucia Pallottino upon evaluation of the Associate Editor and Reviewers' comments. This work is funded in part by the German Research Foundation (DFG): RTG 2236/2 (UnRAVeL). We thank the NHR4CES at RWTH Aachen for computing resources (p0027430, rwth2087). OpenAI Codex assisted with coding.}%
    \thanks{$^1$ Institute for Data Science in Mechanical Engineering (DSME), RWTH Aachen University, Germany
    {\tt\small \{hose.henrik, trimpe\}@dsme.rwth-aachen.de}}%
    \thanks{$^2$ Department of Mechanical Engineering, Massachusetts Institute of Technology, Cambridge MA, USA.}%
    \thanks{Digital Object Identifier (DOI): see top of this page.}%
}
\fi

\markboth{IEEE Robotics and Automation Letters. Preprint Version. Accepted November, 2026}%
{Hose \MakeLowercase{\textit{et al.}}: WarpMPC: Large-Batch MPC on GPU}

\maketitle

\begin{abstract}
This paper introduces numerical optimizations for maximizing throughput on GPU when solving large batches (10\,000 to over 100\,000) of sequential quadratic programming (SQP) iterations, where all problems have the same structure.
The optimizations are implemented in a toolbox called \toolboxname for model-predictive control (MPC) in JAX and Warp.
Based on the insight that all MPC problem instances in a batch share the same sparsity in time, cost, and constraints, we propose unrolling sparse linear factorizations and solves, which dominate alternating direction method of multipliers (ADMM) solver runtime.
We avoid memory access bottlenecks and unnecessary computations via optimized memory layout, padding-reducing segmentation of the unrolled factorization, and dependency level scheduled backsolves, additionally accelerating sensitivity computation.
We achieve throughputs of 8\,000 to 250\,000 SQP iterations per second on nonlinear cartpole, quadrotor, and humanoid robot benchmarks, outperforming baselines by 3$\times$ to 25$\times$.
We illustrate practical usefulness by synthesizing a dataset and training a neural network approximation of an MPC in under 5 minutes that stabilizes a nano quadrotor in hardware experiments.
\end{abstract}


\section{Introduction}
\label{sec:introduction}
\IEEEPARstart{M}{odel}-predictive control (MPC) is a popular method for controlling nonlinear systems subject to constraints~\cite{rawlings2017mpc}.
Recently, there has been significant interest in integrating deep learning and MPC for robotics, which requires solving large batches of MPC problems, e.g., in reinforcement learning~\cite{romero2024actor,jeon2025residual},
or imitation learning~\cite{nubert2020safe,hose2024parameter}.
Yet, most MPC toolboxes and solvers are optimized for single solve speed on CPU~\cite{verschueren2022acados,stellato2018osqp} or GPU~\cite{schubiger2020gpu,cole2022exploiting,adabag2024mpcgpu, pacaud2024gpu,kang2024fast}, %
or if batched on GPU cannot handle general constraints~\cite{amos2018differentiable,frostig2021trajax,adabag2024mpcgpu,adabag2026differentiable,amatucci2025primal}.
Therefore, learning applications resort to solving batches of MPC problems sequentially on CPU, which can take days or weeks~\cite{jenelten2024dtc,romero2024actor}, or naively parallelize across many workers, which requires expensive infrastructure like high-performance computing clusters~\cite{hose2024parameter,julbe2025diffusion}.
We propose a solver that achieves high throughput when solving large batches of MPC problems on a single GPU, which enables MPC in learning pipelines.

Nonlinear MPC solves a numerical optimization over a finite prediction horizon at every control loop iteration.
This can be done with sequential quadratic programming (SQP)~\cite{nocedal2006numerical}, which iteratively computes step directions by approximating the true nonlinear problem at the current iterate with a quadratic cost and linear constraints, which makes each step a quadratic program (QP).
Resulting QPs are large, but very sparse ($<$\SI{1}{\percent} structural nonzeros in the Karush-Kuhn-Tucker (KKT) system, see Fig.~\ref{fig:sparsity_patterns}) and can be efficiently solved with Alternating Direction Method of Multipliers (ADMM)~\cite{stellato2018osqp,schubiger2020gpu,adabag2024mpcgpu,adabag2026differentiable,du2025gato} and classic, sparsity exploiting~$LDL^\top$ factorizations~\cite{stewart2003building}.
ADMM is particularly appealing, because every QP solve requires only one sparse factorization, followed by multiple backsolves.
However, sparse linear GPU solvers outperforming CPU are rare (cf.~\cite{nvidia_cudss_preview, swirydowicz2022linear}), because GPUs struggle with branching, variable length loops, and irregular memory access.
Hence, single-solve frameworks use iterative conjugate gradient (CG) methods that allow for parallelization~\cite{schubiger2020gpu,adabag2024mpcgpu, bravopalacios2026turbompc} at the expense of accuracy and more arithmetic computations that limit throughput for large batches.

In this work, we present \toolboxname, a JAX toolbox with Warp backend for solving large batches of MPC problems in parallel using SQP and a filter line search (over \num{100000} parallel solves on a single GPU) with a custom implementation of an operator splitting ADMM solver closely resembling OSQP~\cite{stellato2018osqp,schubiger2020gpu}.
We avoid branching with problem-specific sparse~$LDL^\top$ kernels that are fast on GPU by unrolling the factorization and backsolves (cf. \cite{jeon2024cusadi}, which uses CasADi's $LDL^\top$).
SQP data is computed sparsity-preserving and stage-wise with code automatically exported from CasADi.
We propose several optimizations that yield additional throughput gains: batched memory layout, segmentation of subsequent columns in the~$LDL^\top$ factorization into individual kernels to minimize padding, and parallelization across dependency levels in the backsolve.
With these optimizations, we achieve superior SQP throughputs (\num{8 000} SQP iterations per second even on challenging problems such as whole-body MPC for humanoid locomotion).
The additional factorization and backsolves for computing sensitivities (gradients) of the SQP step likewise benefit from the optimizations, making our toolbox attractive for machine learning applications with differentiable MPCs~\cite{amos2018differentiable,adabag2026differentiable}.
We make the following contributions:
\begin{compactenum}
    \item \toolboxname, an open-source JAX toolbox for solving large batches of MPCs in parallel using SQP and line search.
    \item A custom CasADi to JAX translation for sparsity-preserving dynamics and cost implementation.
    \item An ADMM and sparse~$LDL^\top$ implementation that unrolls the factorization and backsolves into custom kernels for fixed sparsity patterns of the MPC problem.
    \item Memory layout optimization, segmented kernels, and backsolve dependency level scheduled parallelization that significantly improve throughput of the unrolling.
\end{compactenum}
We extensively benchmark the contributions against state-of-the-art libraries on GPU showing consistent throughput improvements of $3\times$ to $25\times$ for large batches, including three realistic, nonlinear MPCs for a cartpole, quadrotor, and humanoid robot.
We illustrate practical relevance by distilling neural network approximations from an MPC in under \SI{5}{\minute} that stabilize a nano quadrotor in hardware experiments.
\rev{We further show pingpong juggling on a Franka robot as contact-implicit experiment and provide a motion retargeting example in the supplementary code.}
\ifanonymized
WarpMPC's code is available in the supplementary materials and a video of our experiments is available at: \href{https://youtu.be/ltVbiddqG_c}{youtu.be/ltVbiddqG\_c}
\else
WarpMPC is published at \href{https://github.com/hshose/warpmpc}{github.com/hshose/warpmpc} and a video of our experiments is available at: \href{https://youtu.be/bZhwcS61y24}{youtu.be/bZhwcS61y24}
\fi

\subsection{Related Work}
Many CPU based MPC toolboxes exist (e.g.,~\cite{Andersson2019,verschueren2022acados,hart2017pyomo,fiedler2023mpc}), some even support batched solves~\cite{verschueren2022acados}.
Yet, direct integration with GPU machine learning frameworks is limited (cf.~\cite{salzmann2024learning,lahr2026l4acados}) and batch sizes are hundreds of parallel solves at most.

General purpose batched dense QP solvers for GPU exist~(e.g., \cite{tracy2024differentiability}), but sparse implementations for GPU are rare (cf. MPAX~\cite{lu2025mpax}).
Interior-point method nonlinear solvers can be accelerated with GPUs~\cite{pacaud2024gpu,regev2023hykkt}, but are difficult to warm start and require frequent factorizations compared to ADMM.
SQP with ADMM and iterative LQR (iLQR) methods are popular choices for MPC specific solvers on GPU, as summarized in Table~\ref{tab:relatedwork}.
ADMM solvers introduced in~\cite{adabag2024mpcgpu,adabag2026differentiable,du2025gato} and iLQR~\cite{amatucci2025primal,amos2018differentiable,frostig2021trajax} are fast on GPU, but cannot handle hard inequality constraints that are essential for many applications. \cite{amatucci2025primal} implements a soft barrier penalty for inequality constraints.
OSQP~\cite{stellato2018osqp} style ADMM with operator splitting capable of general inequality constraints in the QP can be implemented on GPU~\cite{schubiger2020gpu}, even for MPC~\cite{bishop2024relu,jeon2025residual,bravopalacios2026turbompc}.
In the case of~\cite{bishop2024relu}, a dense reformulation is used with a custom linear solve that provides great speed ups due to pre-computed factorization, but requires fixed cost Hessian and constraint Jacobian and therefore does not apply to SQP.
In \cite{jeon2025residual, bravopalacios2026turbompc}, NVIDIA cuDSS sparse linear algebra routines are used, which are fast for small batches, but are outperformed on large batches by the optimizations proposed in this paper.
Further, \cite{jeon2025residual} does not compute sensitivities, which we do.

\newcommand{\yes}{\textbf{Yes}}
\begin{table}[htb]\vspace{-1em}
    \centering
    \caption{GPU Solvers for Trajectory Optimization and MPC}
    \vspace{-0.5em}
    \label{tab:relatedwork}
    \resizebox{3.4in}{!}{%
    \setlength{\tabcolsep}{3pt}
    \begin{tabular}{@{}lcccccc@{}}
        \toprule
        \textbf{Paper}
        & \makecell{\textbf{Outer}\\\textbf{Alg.}}
        & \makecell{\textbf{Linear}\\\textbf{Solver}}
        & \makecell{\textbf{Hard Ineq.}\\\textbf{Constraints}}
        & \makecell{\textbf{Sensiti-}\\\textbf{vities}}
        & \makecell{\textbf{Horizon}\\\textbf{Sparsity}}
        & \makecell{\textbf{Dynamics}\\\textbf{Sparsity}}\\
        \midrule
        \toolboxname (ours)
        & SQP with ADMM
        & Unrolled sparse~$LDL^\top$
        & \yes
        & \yes
        & \yes
        & \yes \\
        \midrule
        TurboMPC~\cite{bravopalacios2026turbompc}
        & SQP with ADMM
        & cuDSS~$LDL^\top$, PCG
        & \yes
        & \yes
        & \yes
        & No \\
        Jeon et al. \cite{jeon2025residual}
        & SQP with ADMM
        & cuDSS~$LDL^\top$
        & \yes
        & No
        & \yes
        & \yes \\
        DiffMPC~\cite{adabag2026differentiable}
        & SQP with ADMM
        & PCG
        & No
        & \yes
        & \yes
        & No \\
        Kang et al. \cite{kang2024fast}
        & SDP with ADMM
        & CPU~$LDL^\top$
        & No
        & \yes
        & \yes 
        & No
        \\
        ReLU-QP \cite{bishop2024relu}
        & Linear MPC, ADMM
        & Precomputed
        & \yes
        & No
        & No
        & No\\
        MPCGPU~\cite{adabag2024mpcgpu,du2025gato}
        & Direct TO
        & PCG
        & No
        & No
        & \yes
        & No \\
        MPX \cite{amatucci2025primal}
        & iLQR, DDP
        & Parallel scan
        & No
        & No
        & \yes
        & \yes\\
        mpc.pytorch~\cite{amos2018differentiable}
        & iLQR
        & Riccati recursion
        & No
        & \yes
        & \yes
        & No\\
        Trajax \cite{frostig2021trajax}
        & iLQR
        & Riccati recursion
        & No
        & \yes
        & \yes
        & No \\
        \bottomrule
        \end{tabular}%
  }
  \vspace{-1em}
\end{table}

\section{Background}
\label{sec:background}

\subsection{Nonlinear Model-Predictive Control}
\label{subsec:mpc_background}

We write a nonlinear model-predictive control (MPC) problem over a horizon of
$N$ intervals in stage form.  Let $z_k\in\R^{n_k}$ denote the decision variables
at stage $k$, typically containing state and input components, and let
$p_k\in\R^{r_k}$ collect stage parameters such as references, measured initial
state, time step, or mode data.  A generic sparse stage formulation is
\begin{equation}
\begin{array}{ll}
    \underset{z_0,\ldots,z_N}{\operatorname{minimize}} &
    \displaystyle
    \sum_{k=0}^{N-1}\ell_k(z_k,z_{k+1},p_k) + \ell_N(z_N,p_N) \\
    \operatorname{subject\ to} &
    l_k(p_k)\leq g_k(z_k,z_{k+1},p_k)\leq u_k(p_k),\\
    & k=0,\ldots,N-1,\\
    & l_N(p_N)\leq g_N(z_N,p_N)\leq u_N(p_N).
\end{array}
\label{eq:nmpc_stage_form}
\end{equation}
This notation includes the usual dynamics constraints by placing
$x_{k+1}-f_k(x_k,u_k,p_k)$ in $g_k\in\mathbb{R}^{m_k}$ with equal lower and upper bounds.  Initial state constraints, path constraints, control limits, and terminal constraints are represented by additional components of $g_k$ and by their corresponding bounds.  The important structural property is locality: nonterminal stage functions couple only neighboring decision blocks $(z_k,z_{k+1})$.
Consequently, the Jacobian of all constraints is block banded and the Hessian of the objective has nonzeros only inside the same neighboring stage blocks (see Fig.~\ref{fig:sparsity_patterns}), which many solvers exploit (Tab.~\ref{tab:relatedwork}, Horizon Sparsity).
Additionally, many MPC formulations show sparsity within a single block (callouts in Fig.~\ref{fig:sparsity_patterns}), which is less commonly exploited (Tab.~\ref{tab:relatedwork}, Dynamics Sparsity).
For high-dimensional systems and long-horizons typical in robotics, sparsity exploitation becomes advantageous compared to dense reformulations~\cite{rawlings2017mpc}.

\newcommand{\KktCartpoleSparsityImage}{figures/kkt_sparsity_cartpole.png}
\newcommand{\KktCartpoleHessianCalloutImage}{figures/kkt_callout_cartpole_hessian_t50.png}
\newcommand{\KktCartpoleJacobianCalloutImage}{figures/kkt_callout_cartpole_constraint_jacobian_t50.png}
\newcommand{\KktCrazyflieSparsityImage}{figures/kkt_sparsity_crazyflie.png}
\newcommand{\KktCrazyflieHessianCalloutImage}{figures/kkt_callout_crazyflie_hessian_t20.png}
\newcommand{\KktCrazyflieJacobianCalloutImage}{figures/kkt_callout_crazyflie_constraint_jacobian_t20.png}
\newcommand{\KktHumanoidSparsityImage}{figures/kkt_sparsity_humanoid.png}
\newcommand{\KktHumanoidHessianCalloutImage}{figures/kkt_callout_humanoid_hessian_t13.png}
\newcommand{\KktHumanoidJacobianCalloutImage}{figures/kkt_callout_humanoid_constraint_jacobian_t13.png}

\newcommand{\KktCartpoleDimPx}{1111}
\newcommand{\KktCartpoleKktDimPx}{1111}
\newcommand{\KktCartpoleKktNnz}{4723}
\newcommand{\KktCartpoleHessianRowsPx}{505}
\newcommand{\KktCartpoleHessianColsPx}{505}
\newcommand{\KktCartpoleHessianNnz}{505}
\newcommand{\KktCartpoleJacobianRowsPx}{606}
\newcommand{\KktCartpoleJacobianColsPx}{505}
\newcommand{\KktCartpoleJacobianNnz}{1806}
\newcommand{\KktCartpoleJacobianKktNnz}{3612}
\newcommand{\KktCartpoleConstraintDiagonalDimPx}{606}
\newcommand{\KktCartpoleConstraintDiagonalNnz}{606}
\newcommand{\KktCartpoleTimestep}{50}
\newcommand{\KktCartpoleHessianCalloutWidthPx}{10}
\newcommand{\KktCartpoleHessianCalloutHeightPx}{10}
\newcommand{\KktCartpoleJacobianCalloutWidthPx}{10}
\newcommand{\KktCartpoleJacobianCalloutHeightPx}{6}

\newcommand{\KktCartpoleHessianXZeroBL}{250}
\newcommand{\KktCartpoleHessianYZeroBL}{851}
\newcommand{\KktCartpoleHessianXOneBL}{260}
\newcommand{\KktCartpoleHessianYOneBL}{861}

\newcommand{\KktCartpoleJacobianXZeroBL}{250}
\newcommand{\KktCartpoleJacobianYZeroBL}{296}
\newcommand{\KktCartpoleJacobianXOneBL}{260}
\newcommand{\KktCartpoleJacobianYOneBL}{302}

\newcommand{\KktCartpoleJacobianTransposeXZeroBL}{809}
\newcommand{\KktCartpoleJacobianTransposeYZeroBL}{851}
\newcommand{\KktCartpoleJacobianTransposeXOneBL}{815}
\newcommand{\KktCartpoleJacobianTransposeYOneBL}{861}

\newcommand{\KktCrazyflieDimPx}{1312}
\newcommand{\KktCrazyflieKktDimPx}{1312}
\newcommand{\KktCrazyflieKktNnz}{7104}
\newcommand{\KktCrazyflieHessianRowsPx}{656}
\newcommand{\KktCrazyflieHessianColsPx}{656}
\newcommand{\KktCrazyflieHessianNnz}{656}
\newcommand{\KktCrazyflieJacobianRowsPx}{656}
\newcommand{\KktCrazyflieJacobianColsPx}{656}
\newcommand{\KktCrazyflieJacobianNnz}{2896}
\newcommand{\KktCrazyflieJacobianKktNnz}{5792}
\newcommand{\KktCrazyflieConstraintDiagonalDimPx}{656}
\newcommand{\KktCrazyflieConstraintDiagonalNnz}{656}
\newcommand{\KktCrazyflieTimestep}{20}
\newcommand{\KktCrazyflieHessianCalloutWidthPx}{32}
\newcommand{\KktCrazyflieHessianCalloutHeightPx}{32}
\newcommand{\KktCrazyflieJacobianCalloutWidthPx}{32}
\newcommand{\KktCrazyflieJacobianCalloutHeightPx}{16}

\newcommand{\KktCrazyflieHessianXZeroBL}{320}
\newcommand{\KktCrazyflieHessianYZeroBL}{960}
\newcommand{\KktCrazyflieHessianXOneBL}{352}
\newcommand{\KktCrazyflieHessianYOneBL}{992}

\newcommand{\KktCrazyflieJacobianXZeroBL}{320}
\newcommand{\KktCrazyflieJacobianYZeroBL}{308}
\newcommand{\KktCrazyflieJacobianXOneBL}{352}
\newcommand{\KktCrazyflieJacobianYOneBL}{324}

\newcommand{\KktCrazyflieJacobianTransposeXZeroBL}{988}
\newcommand{\KktCrazyflieJacobianTransposeYZeroBL}{960}
\newcommand{\KktCrazyflieJacobianTransposeXOneBL}{1004}
\newcommand{\KktCrazyflieJacobianTransposeYOneBL}{992}

\newcommand{\KktHumanoidDimPx}{4938}
\newcommand{\KktHumanoidKktDimPx}{4938}
\newcommand{\KktHumanoidKktNnz}{70810}
\newcommand{\KktHumanoidHessianRowsPx}{1704}
\newcommand{\KktHumanoidHessianColsPx}{1704}
\newcommand{\KktHumanoidHessianNnz}{2676}
\newcommand{\KktHumanoidJacobianRowsPx}{3234}
\newcommand{\KktHumanoidJacobianColsPx}{1704}
\newcommand{\KktHumanoidJacobianNnz}{32450}
\newcommand{\KktHumanoidJacobianKktNnz}{64900}
\newcommand{\KktHumanoidConstraintDiagonalDimPx}{3234}
\newcommand{\KktHumanoidConstraintDiagonalNnz}{3234}
\newcommand{\KktHumanoidTimestep}{13}
\newcommand{\KktHumanoidHessianCalloutWidthPx}{120}
\newcommand{\KktHumanoidHessianCalloutHeightPx}{120}
\newcommand{\KktHumanoidJacobianCalloutWidthPx}{120}
\newcommand{\KktHumanoidJacobianCalloutHeightPx}{114}

\newcommand{\KktHumanoidHessianXZeroBL}{816}
\newcommand{\KktHumanoidHessianYZeroBL}{4002}
\newcommand{\KktHumanoidHessianXOneBL}{936}
\newcommand{\KktHumanoidHessianYOneBL}{4122}

\newcommand{\KktHumanoidJacobianXZeroBL}{816}
\newcommand{\KktHumanoidJacobianYZeroBL}{1550}
\newcommand{\KktHumanoidJacobianXOneBL}{936}
\newcommand{\KktHumanoidJacobianYOneBL}{1664}

\newcommand{\KktHumanoidJacobianTransposeXZeroBL}{3274}
\newcommand{\KktHumanoidJacobianTransposeYZeroBL}{4002}
\newcommand{\KktHumanoidJacobianTransposeXOneBL}{3388}
\newcommand{\KktHumanoidJacobianTransposeYOneBL}{4122}

\DeclareRobustCommand{\KktFullSystemSymbol}{%
    \tikz[baseline=0pt, x=0.3em, y=0.3em]{%
        \fill[RoyalBlue!30] (0,1) rectangle (1,2);
        \fill[BurntOrange!35] (1,1) rectangle (2,2);
        \fill[BurntOrange!35] (0,0) rectangle (1,1);
        \fill[gray!35] (1,0) rectangle (2,1);
        \draw[black, line width=0.3pt] (0,0) rectangle (2,2);
    }%
}
\DeclareRobustCommand{\KktHessianCalloutSymbol}{%
    \tikz[baseline=0pt, x=0.5em, y=0.5em]{%
        \fill[RoyalBlue!30] (0,0) rectangle (1,1);
        \draw[RoyalBlue, line width=0.4pt] (0,0) rectangle (1,1);
    }%
}
\DeclareRobustCommand{\KktJacobianCalloutSymbol}{%
    \tikz[baseline=0pt, x=0.6em, y=0.6em]{%
        \fill[BurntOrange!35] (0,0) rectangle (1,1);
        \draw[BurntOrange, line width=0.4pt] (0,0) rectangle (1,1);
    }%
}
\DeclareRobustCommand{\KktLowerRightCalloutSymbol}{%
    \tikz[baseline=0pt, x=0.6em, y=0.6em]{%
        \fill[gray!35] (0,0) rectangle (1,1);
        \draw[gray, line width=0.4pt] (0,0) rectangle (1,1);
    }%
}
\begin{figure}[t]
    \centering
    \newlength{\kktSparsityHeight}
    \newlength{\kktCalloutGap}
    \newlength{\kktPanelGap}
    \newlength{\kktCalloutHeight}
    \newlength{\kktStatsWidth}
    \newlength{\kktStatsNaturalWidth}
    \setlength{\kktSparsityHeight}{0.68in}
    \setlength{\kktCalloutGap}{0.025in}
    \setlength{\kktPanelGap}{0.06in}
    \setlength{\kktCalloutHeight}{\dimexpr\kktSparsityHeight/2-\kktCalloutGap/2\relax}
    \setlength{\kktStatsWidth}{\dimexpr\kktSparsityHeight+\kktCalloutGap+\kktCalloutHeight\relax}
    \setlength{\kktStatsNaturalWidth}{2in}
    \newcommand{\KktSystemStats}[2]{%
        \resizebox{\kktStatsWidth}{!}{%
            \begin{minipage}{\kktStatsNaturalWidth}
                \centering\normalsize
                #1\\[-0.2ex]
                #2
            \end{minipage}%
        }%
    }
    \pgfmathtruncatemacro{\KktCartpoleHorizonLength}{2*\KktCartpoleHessianColsPx/\KktCartpoleHessianCalloutWidthPx - 1}
    \pgfmathtruncatemacro{\KktCrazyflieHorizonLength}{2*\KktCrazyflieHessianColsPx/\KktCrazyflieHessianCalloutWidthPx - 1}
    \pgfmathtruncatemacro{\KktHumanoidHorizonLength}{2*\KktHumanoidHessianColsPx/\KktHumanoidHessianCalloutWidthPx - 1}
    \edef\KktCartpoleKktNnzPercent{\fpeval{round(100*\KktCartpoleKktNnz/(\KktCartpoleKktDimPx*\KktCartpoleKktDimPx),2)}}
    \edef\KktCrazyflieKktNnzPercent{\fpeval{round(100*\KktCrazyflieKktNnz/(\KktCrazyflieKktDimPx*\KktCrazyflieKktDimPx),2)}}
    \edef\KktHumanoidKktNnzPercent{\fpeval{round(100*\KktHumanoidKktNnz/(\KktHumanoidKktDimPx*\KktHumanoidKktDimPx),2)}}
    \begin{tikzpicture}[
        every node/.style={inner sep=0pt, outer sep=0pt},
        kkt box/.style={draw=black, line width=0.4pt},
        hessian box/.style={draw=RoyalBlue, line width=0.4pt},
        jacobian box/.style={draw=BurntOrange, line width=0.4pt},
        system title/.style={align=center, font=\scriptsize}
    ]
        \node[anchor=south west] (cpkkt) at (0,0)
            {\includegraphics[height=\kktSparsityHeight]{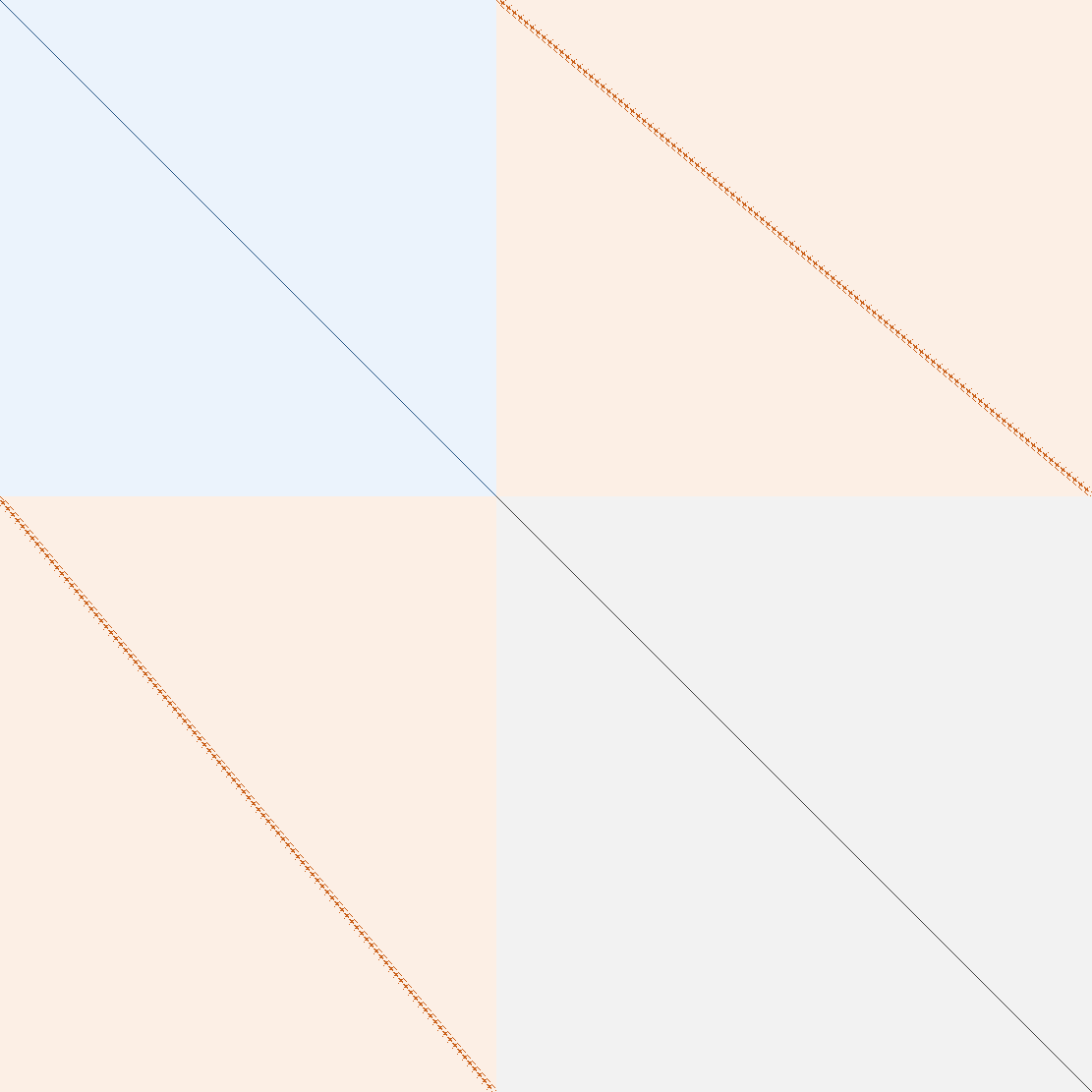}};
        \draw[kkt box] (cpkkt.south west) rectangle (cpkkt.north east);
        \begin{scope}[shift={(cpkkt.south west)}, x=\kktSparsityHeight, y=\kktSparsityHeight]
            \draw[hessian box]
                ({\KktCartpoleHessianXZeroBL/\KktCartpoleDimPx}, {\KktCartpoleHessianYZeroBL/\KktCartpoleDimPx})
                rectangle
                ({\KktCartpoleHessianXOneBL/\KktCartpoleDimPx}, {\KktCartpoleHessianYOneBL/\KktCartpoleDimPx});
            \draw[jacobian box]
                ({\KktCartpoleJacobianXZeroBL/\KktCartpoleDimPx}, {\KktCartpoleJacobianYZeroBL/\KktCartpoleDimPx})
                rectangle
                ({\KktCartpoleJacobianXOneBL/\KktCartpoleDimPx}, {\KktCartpoleJacobianYOneBL/\KktCartpoleDimPx});
            \draw[jacobian box]
                ({\KktCartpoleJacobianTransposeXZeroBL/\KktCartpoleDimPx}, {\KktCartpoleJacobianTransposeYZeroBL/\KktCartpoleDimPx})
                rectangle
                ({\KktCartpoleJacobianTransposeXOneBL/\KktCartpoleDimPx}, {\KktCartpoleJacobianTransposeYOneBL/\KktCartpoleDimPx});
        \end{scope}
        \node[anchor=north west, xshift=\kktCalloutGap] (cphessian) at (cpkkt.north east)
            {\includegraphics[height=\kktCalloutHeight]{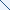}};
        \node[anchor=south west, xshift=\kktCalloutGap] (cpjacobian) at (cpkkt.south east)
            {\reflectbox{\rotatebox{-90}{\includegraphics[width=\kktCalloutHeight]{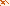}}}};
        \draw[hessian box] (cphessian.south west) rectangle (cphessian.north east);
        \draw[jacobian box] (cpjacobian.south west) rectangle (cpjacobian.north east);

        \node[anchor=south west, xshift=\kktPanelGap] (cfkkt) at (cphessian.south east |- cpkkt.south west)
            {\includegraphics[height=\kktSparsityHeight]{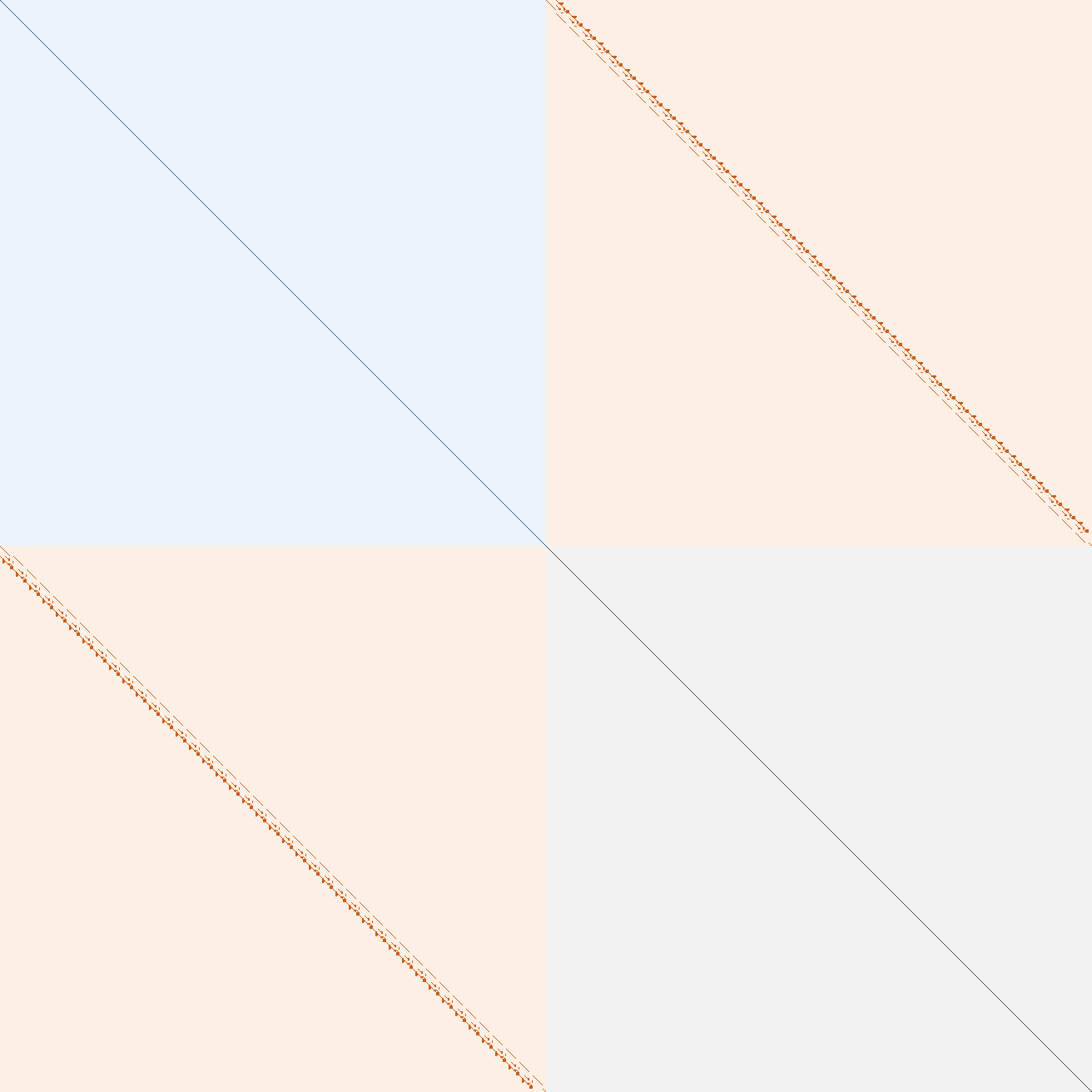}};
        \draw[kkt box] (cfkkt.south west) rectangle (cfkkt.north east);
        \begin{scope}[shift={(cfkkt.south west)}, x=\kktSparsityHeight, y=\kktSparsityHeight]
            \draw[hessian box]
                ({\KktCrazyflieHessianXZeroBL/\KktCrazyflieDimPx}, {\KktCrazyflieHessianYZeroBL/\KktCrazyflieDimPx})
                rectangle
                ({\KktCrazyflieHessianXOneBL/\KktCrazyflieDimPx}, {\KktCrazyflieHessianYOneBL/\KktCrazyflieDimPx});
            \draw[jacobian box]
                ({\KktCrazyflieJacobianXZeroBL/\KktCrazyflieDimPx}, {\KktCrazyflieJacobianYZeroBL/\KktCrazyflieDimPx})
                rectangle
                ({\KktCrazyflieJacobianXOneBL/\KktCrazyflieDimPx}, {\KktCrazyflieJacobianYOneBL/\KktCrazyflieDimPx});
            \draw[jacobian box]
                ({\KktCrazyflieJacobianTransposeXZeroBL/\KktCrazyflieDimPx}, {\KktCrazyflieJacobianTransposeYZeroBL/\KktCrazyflieDimPx})
                rectangle
                ({\KktCrazyflieJacobianTransposeXOneBL/\KktCrazyflieDimPx}, {\KktCrazyflieJacobianTransposeYOneBL/\KktCrazyflieDimPx});
        \end{scope}
        \node[anchor=north west, xshift=\kktCalloutGap] (cfhessian) at (cfkkt.north east)
            {\includegraphics[height=\kktCalloutHeight]{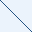}};
        \node[anchor=south west, xshift=\kktCalloutGap] (cfjacobian) at (cfkkt.south east)
            {\reflectbox{\rotatebox{-90}{\includegraphics[width=\kktCalloutHeight]{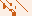}}}};
        \draw[hessian box] (cfhessian.south west) rectangle (cfhessian.north east);
        \draw[jacobian box] (cfjacobian.south west) rectangle (cfjacobian.north east);

        \node[anchor=south west, xshift=\kktPanelGap] (humkkt) at (cfhessian.south east |- cfkkt.south west)
            {\includegraphics[height=\kktSparsityHeight]{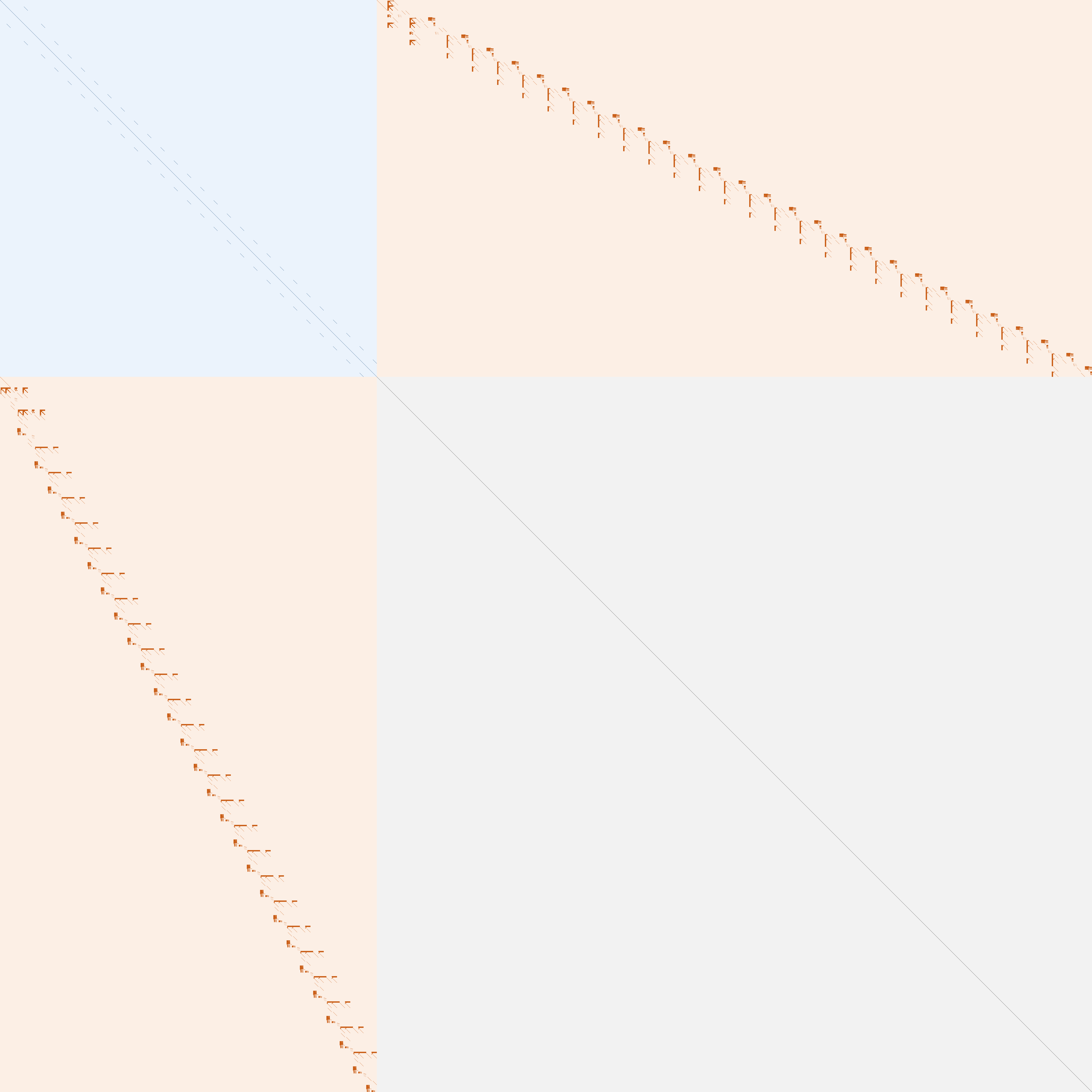}};
        \draw[kkt box] (humkkt.south west) rectangle (humkkt.north east);
        \begin{scope}[shift={(humkkt.south west)}, x=\kktSparsityHeight, y=\kktSparsityHeight]
            \draw[hessian box]
                ({\KktHumanoidHessianXZeroBL/\KktHumanoidDimPx}, {\KktHumanoidHessianYZeroBL/\KktHumanoidDimPx})
                rectangle
                ({\KktHumanoidHessianXOneBL/\KktHumanoidDimPx}, {\KktHumanoidHessianYOneBL/\KktHumanoidDimPx});
            \draw[jacobian box]
                ({\KktHumanoidJacobianXZeroBL/\KktHumanoidDimPx}, {\KktHumanoidJacobianYZeroBL/\KktHumanoidDimPx})
                rectangle
                ({\KktHumanoidJacobianXOneBL/\KktHumanoidDimPx}, {\KktHumanoidJacobianYOneBL/\KktHumanoidDimPx});
            \draw[jacobian box]
                ({\KktHumanoidJacobianTransposeXZeroBL/\KktHumanoidDimPx}, {\KktHumanoidJacobianTransposeYZeroBL/\KktHumanoidDimPx})
                rectangle
                ({\KktHumanoidJacobianTransposeXOneBL/\KktHumanoidDimPx}, {\KktHumanoidJacobianTransposeYOneBL/\KktHumanoidDimPx});
        \end{scope}
        \node[anchor=north west, xshift=\kktCalloutGap] (humhessian) at (humkkt.north east)
            {\includegraphics[height=\kktCalloutHeight]{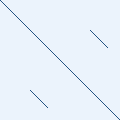}};
        \node[anchor=south west, xshift=\kktCalloutGap] (humjacobian) at (humkkt.south east)
            {\reflectbox{\rotatebox{-90}{\includegraphics[width=\kktCalloutHeight]{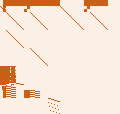}}}};
        \draw[hessian box] (humhessian.south west) rectangle (humhessian.north east);
        \draw[jacobian box] (humjacobian.south west) rectangle (humjacobian.north east);

        \node[system title, anchor=south, yshift=0.025in] at ($(cpkkt.north west)!0.5!(cphessian.north east)$)
            {\textbf{Cartpole}\\
            \KktSystemStats
                {$N=\KktCartpoleHorizonLength$, $n=\KktCartpoleHessianColsPx$, $m=\KktCartpoleJacobianRowsPx$}
                {$\nnz(\K)=\KktCartpoleKktNnz$ (\KktCartpoleKktNnzPercent\%)}};
        \node[system title, anchor=south, yshift=0.025in] at ($(cfkkt.north west)!0.5!(cfhessian.north east)$)
            {\textbf{Quadrotor}\\
            \KktSystemStats
                {$N=\KktCrazyflieHorizonLength$, $n=\KktCrazyflieHessianColsPx$, $m=\KktCrazyflieJacobianRowsPx$}
                {$\nnz(\K)=\KktCrazyflieKktNnz$ (\KktCrazyflieKktNnzPercent\%)}};
        \node[system title, anchor=south, yshift=0.025in] at ($(humkkt.north west)!0.5!(humhessian.north east)$)
            {\textbf{Humanoid}\\
            \KktSystemStats
                {$N=\KktHumanoidHorizonLength$, $n=\KktHumanoidHessianColsPx$, $m=\KktHumanoidJacobianRowsPx$}
                {$\nnz(\K)=\KktHumanoidKktNnz$ (\KktHumanoidKktNnzPercent\%)}};
    \end{tikzpicture}
    \vspace{-1em}
    \caption{Sparsity patterns of the full KKT systems $\K$ (\KktFullSystemSymbol) that are factorized once per SQP iteration. Shown here are the cartpole, quadrotor, and humanoid robot. Dark pixels indicate nonzero entries. The cost Hessian $H$ blocks (\KktHessianCalloutSymbol) and constraint Jacobian $J_g$ blocks (\KktJacobianCalloutSymbol) show the typical block-banded sparsity pattern that enables parallel linearization at the current SQP iterate. Callouts are a neighboring timesteps ($z_k, z_{k+1}$) block.}
    \vspace{-1.5em}
    \label{fig:sparsity_patterns}
\end{figure}

\subsection{Sequential Quadratic Programming with Line Search}
\label{subsec:sqp_background}

Let $z\in\mathbb{R}^n$ stack all stage variables in \eqref{eq:nmpc_stage_form}, $g(z)\in\mathbb{R}^m$ stack all constraints, and write the NLP compactly as
\begin{equation}
    \underset{z}{\operatorname{minimize}}\ f(z)
    \quad\text{subject\ to}\quad
    l\leq g(z)\leq u .
\label{eq:nlp_compact}
\end{equation}
SQP computes a step $d^j\in\mathbb{R}^n$ at iterate $z^j\in\mathbb{R}^n$ from a quadratic approximation of the objective and a first-order approximation of the constraints,
\begin{equation}
\begin{array}{ll}
    \underset{d}{\operatorname{minimize}} &
    \frac{1}{2}d^\top H^j d + \nabla f(z^j)^\top d\\
    \operatorname{subject\ to} &
    l-g(z^j) \leq J_g(z^j)d \leq u-g(z^j),
\end{array}
\label{eq:sqp_qp}
\end{equation}
where $J_g(z^j)=\partial g / \partial z(z^j)\in\mathbb{R}^{m \times n}$ is the constraint Jacobian and $H^j\in\mathbb{R}^{n\times n}$ is a symmetric positive semidefinite Gauss-Newton Hessian approximation of $\partial^2 f / \partial z^2 (z^j)$.
\rev{In our benchmarks, costs are quadratic and the resulting Hessians positive definite. We do not implement automatic construction of Gauss-Newton Hessian approximations of arbitrary costs.}

The full SQP step $z^j+d^j$ can be too aggressive for nonlinear constraints.
Therefore, a damped update of the form
\begin{equation}
    z^{j+1}=z^j+\alpha^j d^j,\qquad \alpha^j\in(0,1],
\label{eq:sqp_update}
\end{equation}
is common, where $\alpha$ is selected from a finite geometric ladder.
Algorithms for selecting $\alpha^j$ include filter line searches~\cite{wachter2006implementation}, which we implement without second-order corrections or backtracking, and backtracking on merit functions~\cite{nocedal2006numerical}.

\subsection{Operator Splitting Direct ADMM Step}
\label{subsec:osqp_background}

Problem~\eqref{eq:sqp_qp} can be compactly written as a convex~QP
\begin{equation}
\begin{array}{ll}
    \underset{\xi\in\R^n}{\operatorname{minimize}}\ 
        \frac{1}{2}\xi^\top P \xi + q^\top \xi \quad
    \operatorname{subject\ to} \quad \underline{b} \leq A \xi \leq \overline{b},
\end{array}
\label{eq:qp}
\end{equation}
where $P\succeq 0$, $A\in\R^{m\times n}$, and the inequalities may include
one-sided or equality constraints.
This QP can be solved by many generic solvers, yet ADMM flavored solvers are particularly appealing in practice due to very few per-iteration computations, excellent warm starting capabilities, sparsity exploitation, and fast convergence to accuracy sufficient for closed-loop control.
We focus on OSQP's~\cite{stellato2018osqp} implementation of operator splitting $s=A\xi$ and $\C=\{s\mid \underline{b}\leq s\leq \overline{b}\}$.
At each iterate $t$, the algorithm solves the quasi-definite linear system
\begin{equation}
\K
\begin{bmatrix}
\tilde{\xi}^{t+1}\\ \nu^{t+1}
\end{bmatrix}
=
\begin{bmatrix}
\sigma \xi^t-q\\ s^t-\Gamma^{-1}y^t
\end{bmatrix},
\quad
\K =
\begin{bmatrix}
P+\sigma I & A^\top\\
A & -\Gamma^{-1}
\end{bmatrix}
\label{eq:kkt}
\end{equation}
with proximal parameter $\sigma>0$, relaxation parameter $\omega_{\rm rel}\in(0,2)$, and positive diagonal $\Gamma=\diag(\rho_1,\ldots,\rho_m)$.
It then updates and projects:
\begin{align}
    \begin{split}
    \tilde{s}^{t+1}&=s^t+\Gamma^{-1}(\nu^{t+1}-y^t),\\
    \xi^{t+1} &= \omega_{\rm rel} \tilde{\xi}^{t+1} + (1-\omega_{\rm rel})\xi^t,\label{eq:osqp}\\
    v^{t+1} &= \omega_{\rm rel} \tilde{s}^{t+1} + (1-\omega_{\rm rel})s^t + \Gamma^{-1}y^t,\\
    s^{t+1} &= \Pi_{\C}(v^{t+1}),\\
    y^{t+1} &= y^t + \Gamma(\omega_{\rm rel}\tilde{s}^{t+1}
    +(1-\omega_{\rm rel})s^t-s^{t+1}).
    \end{split}
\end{align}
Notably, $\K$ is independent of $t$, thus the dominating computational tasks are one factorization and many backsolves.

\subsection{Sensitivities of the QP Solution}
\label{subsec:qp_sensitivities}

An outer objective
$\varphi(\xi^\star,y^\star)\in\mathbb{R}$ may depend on the solution of
\eqref{eq:qp}.  Reverse-mode differentiation computes vector--Jacobian
products of the solution map
$(P,q,A,l,u)\mapsto(\xi^\star,y^\star)$ from cotangents
$\bar{\xi}=\partial\varphi/\partial \xi^\star$ and
$\bar{y}=\partial\varphi/\partial y^\star$, without forming the full Jacobian.
OSQP obtains these derivatives by differentiating the QP optimality conditions~\cite{stellato2018osqp}.
Finite lower and upper bounds are written as inequalities
$G\xi\leq\beta$, and equalities as $E\xi=e$.
The signed OSQP dual is converted to nonnegative inequality multipliers $\lambda$, and inequality slack $\upsilon=G\xi^\star-\beta$.
The adjoint variables are initialized as
\begin{equation}
\begin{gathered}
\underbrace{
\left[\begin{smallmatrix}
(1+\varepsilon)I & \mathcal{M}\\
\mathcal{M}^{\top} & -\varepsilon I
\end{smallmatrix}\right]
}_{\mathcal{S}_{\varepsilon}}
\left[\begin{smallmatrix}
\mu^0\\
\psi^0
\end{smallmatrix}\right]
\!=\!-
b,\quad
\mathcal{M}\!=\!
\left[\begin{smallmatrix}
P & G^\top\!\diag(\lambda) & E^\top\\
G & \diag(\upsilon) & 0\\
E & 0 & 0
\end{smallmatrix}\right],
\end{gathered}
\label{eq:osqp_adjoint_system}
\end{equation}
with $\bar r$ the cotangent vector, $w=[\mu^\top,\psi^\top]^\top$, $b=-[\bar r^\top,0^\top]^\top$ and $\varepsilon>0$ is a small regularization.
Accurate sensitivities require refinement against the unregularized system for $N_\mathrm{ref}$ iterations, $k=0,\ldots,N_{\mathrm{ref}}-1$, which update
$\mathcal{S}_{\varepsilon}\Delta w^k\!=\!b-\mathcal{S}_{0}w^k$,
and $w^{k+1}\!=\!w^k+\Delta w^k$,
reusing the factorization of $\mathcal{S}_{\varepsilon}$ for all correction
solves.
$S_0$ is $S_\varepsilon$ for $\varepsilon=0$.
After the final refinement, let $\psi_\xi$ denote the primal
component of $\psi$, and $r_{\underline{b}}$, $r_{\overline{b}}$ the lower- and
upper-bound adjoints obtained by mapping back to the original OSQP bound rows.
The gradients are then $\nabla_q\varphi\!=\!\psi_\xi$,
$\nabla_{r_{\underline{b}}}\varphi\!=\!r_{\underline{b}}$,
$\nabla_{r_{\overline{b}}}\varphi\!=\!-r_{\overline{b}}$, and with $\psi_i^\xi=(\psi_\xi)_i$,
\begin{align}
    \tfrac{\partial\varphi}{\partial P_{ij}}
    \!=\!\tfrac{1}{2}(\psi_i^\xi\xi^\star_j+\psi_j^\xi\xi^\star_i),\,
    \tfrac{\partial}{\partial A_{ij}}\varphi
    \!=\!y^\star_i\psi_j^\xi+(r_{\overline{b}}-r_{\underline{b}})_i\xi^\star_j
\label{eq:osqp_mat_grads}
\end{align}
with indices $i,j$.
Thus the sensitivity computations require one fixed-pattern factorization of $\mathcal{S}_{\epsilon}$ and many backsolves.

\subsection{Sparse~$LDL^\top$ Factorization}
\label{subsec:qdldl_background}
The matrix $\K$ in \eqref{eq:kkt} and Fig.~\ref{fig:sparsity_patterns} is symmetric quasi-definite because $\sigma I$ makes the~\KktHessianCalloutSymbol\,block positive definite and the~\KktLowerRightCalloutSymbol\,block is negative definite. The same holds for $\mathcal{S}_\varepsilon$ in \eqref{eq:osqp_adjoint_system}.
This matrix can be efficiently factorized with sparse~$LDL^\top$ factorizations~\cite{stewart2003building}.
First, a fill-reducing permutation $Q$ is computed, then the system $\bar{\K}=Q\K Q^\top$ factorized as
$\bar{\K} = L D L^\top$,
where $L$ is unit lower triangular and $D$ is diagonal.
With the permutation, a backsolve with right-hand side $h$ becomes
\begin{equation}
\chi = Q^\top L^{-\top}D^{-1}L^{-1}Qh.
\label{eq:ldl_solve}
\end{equation}
After the first $c-1$ columns, consider the partition
\begin{equation}
\bar{\K}=
\left[\begin{smallmatrix}
\K_{11} & k_{1c} & \K_{31}^\top\\
k_{1c}^\top & \kappa_{cc} & k_{3c}^\top\\
\K_{31} & k_{3c} & \K_{33}
\end{smallmatrix}\right],\quad
L=
\left[\begin{smallmatrix}
L_{11} & 0 & 0\\
L_{c1} & 1 & 0\\
L_{31} & L_{3c} & L_{33}
\end{smallmatrix}\right],
\label{eq:ldl_partition}
\end{equation}
where $\K_{11}\in\mathbb{R}^{(c-1)\times(c-1)}$,
$k_{1c}\in\mathbb{R}^{c-1}$, and
$D=\diag(D_{11},\delta_c,D_{33})$, $D_{11}=\diag(\delta_1,\ldots,\delta_{c-1})$.
Writing out the equations for the $c$-th block column of
$\bar{\K}=LDL^\top$ gives
\begin{align}\label{eq:ldl-colum}
    \begin{bmatrix}
    k_{1c}\\ \kappa_{cc}
    \end{bmatrix}
    &=
    \begin{bmatrix}
        L_{11}D_{11}L_{c1}^\top\\
        L_{c1}D_{11}L_{c1}^\top+\delta_c
    \end{bmatrix},
\end{align}
which can be evaluated in a sparse form: at column $c$ of $\bar{\K}$, $L_{c1}$ and $d_c$ are computed.
With $\delta_c\leftarrow\kappa_{cc}$ and work vector $w$ initialized from the
strict upper entries $\bar{\K}_{ic}$, $i\!<\!c$, the pure arithmetic of the factorization is, for each visited
column~$j\!<\!c$,
\begin{align}
    L_{cj} \leftarrow w_j \delta_j^{-1},\quad \delta_c \leftarrow \delta_c - w_jL_{cj},\label{eq:qdldl_recursion_1}\\
    w_i \leftarrow w_i - L_{ij}w_j,\quad i\!\in j\!+\!1,\dots,c\!-\!1.
\label{eq:qdldl_recursion_2}
\end{align}
Here $w_j=\delta_jL_{cj}$ is the current residual entry of the triangular solve $L_{11}D_{11}L_{c1}^\top=k_{1c}$.
Thus, the first assignment in~\eqref{eq:qdldl_recursion_1} computes one entry of $L_{c1}$, the second accumulates the Schur complement $\delta_c=\kappa_{cc}-L_{c1}D_{11}L_{c1}^\top$, and~\eqref{eq:qdldl_recursion_2} subtracts the contribution of column $j$ from the remaining work entries.

A typical~$LDL^\top$ implementation first computes symbolic data such as the permutation~$Q$, the permuted upper-triangular sparsity pattern, the elimination tree, and the nonzero pattern of $L$.
During numerical factorization, it still performs branching, non-floating-point work such as diagonal lookup, elimination-tree walks, insertion-pointer updates, and variable-length loops over column prefixes.
However, all of these operations depend only on the sparsity pattern of $\K$ and can therefore be precomputed or unrolled, which we do in this paper.
Hence, for factorizations with the same sparsity pattern, the online numerical work reduces to the value-dependent arithmetic in \eqref{eq:qdldl_recursion_1} and \eqref{eq:qdldl_recursion_2} and inverting $D$ in~\eqref{eq:ldl_solve}.

\section{A Fixed-Pattern Batched GPU Solver}
\label{sec:method}

We consider batched SQP iterations for MPC problems that share a stage structure and therefore a QP sparsity pattern. 
Across the batch, only numerical values change: the nonzero locations in $P$, $A$, and the KKT matrix are fixed.
We split the solver into an offline symbolic just-in-time compilation phase and an
online numerical phase, summarized in Algorithm~\ref{alg:offline_online_solver}.

\begin{algorithm}[t]
\caption{WarpMPC's batched SQP solve}
\label{alg:offline_online_solver}
\DontPrintSemicolon
\KwIn{functions $f, l, g, u$, horizon $N$, batched $z$ and $p$}
\textbf{Offline phase (CPU/JIT setup).}\;
Linearize $\ell_k$, $g_k$, translate CasADi to JAX\;
Build sparsity patterns and scatter maps for $P$, $A$\;
Compute infill reducing reordering $Q$\;
Symbolic factorization for fixed-pattern $\bar{K}$\;
Compute optimal segmentation $\mathcal{I}_\kappa$ and level schedule\;
Compile linearization, factorization, solve, line-search\;
\BlankLine
\textbf{Online phase (GPU batched execution).}\;
\For{each SQP iteration}{
    Stagewise $\{\{\nabla\ell_k, \nabla^2\ell_k, J_{g_k}, g_k, l_k, u_k\}_{k=0}^N\}_{b=1}^B$\;
    Scatter into $\{P,q,A,\underline{b},\overline{b}\}_{b=1}^B$\;
    $\bar{\K} \gets$ assemble $\K$ and permute $Q \K Q^\top$\;
    $L, D \gets$ segment-wise factorization of $\bar{\K}$\;
    \For{fixed ADMM iterations}{
        $\tilde{\xi}, \tilde{\nu} \gets$ assemble RHS and backsolve \eqref{eq:kkt}\;
        $\xi, \nu, s, y\gets$ project, relax, dual upd. \eqref{eq:osqp}\;
    }
    $\alpha\gets$ parallel evaluate line-search ladder for $d=\xi$\;
    $z\gets z+\alpha d$, see~\eqref{eq:sqp_update}
}
\end{algorithm}

\subsection{Sparse SQP Data Functions in JAX}
\label{subsec:sparse_sqp_data_kernels}

We implement \eqref{eq:nmpc_stage_form} in a generic, stage-wise form, where each nonterminal stage is a function of $(z_k,z_{k+1},p_k)$ and the terminal stage of $(z_N,p_N)$.
These define the four objects $\ell_k$, $g_k$, $l_k$, $u_k$ without sacrificing flexibility.
At offline construction time, the user provides CasADi functions, from which we compute the Jacobian and Hessian, expand to scalar SX form, and sparsify.
The stage object stores only the fixed triplets needed by SQP: nonzero entries of $\nabla\ell_k$, upper-triangular nonzeros of $\nabla^2\ell_k$, and nonzeros of $J_{g_k}$.
We then automatically translate the CasADi functions to JAX for $(\ell_k,g_k,l_k,u_k)$ and packed structural nonzeros of $\nabla\ell_k$, $\nabla^2\ell_k$, and $J_{g_k}$.
During the online evaluation, we group reused stage functions: typically there is one for the initial stage, one reused for all intermediate stages, and one terminal function.
However, any number of different stages is possible.
Identical stages are evaluated in parallel over stage and batch indices.
From the output of stage functions, the permuted KKT system is assembled. 

\begin{figure}[b]
    \centering
    \vspace{-1em}
    \includegraphics{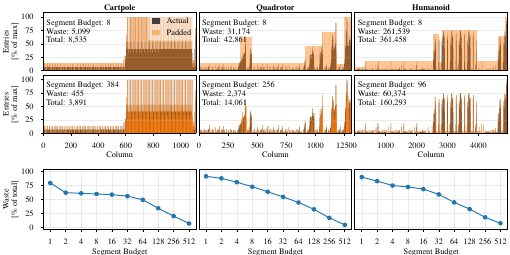}%
    \vspace{-0.5em}
    \caption{Visualization of optimal column segmentation for the cartpole, quadrotor, and humanoid MPC. The number of factor entries per column are plotted as a histogram over the column index for segment budgets of 8 (top) and 384, 256, and 96 (center). Additionally, the percentage of wasted padding entries over segment budget are plotted over segment sizes (bottom).}\label{fig:segmentation}
\end{figure}

\subsection{Optimized Parallel~$LDL^\top$ Factorizations and Solves}
\label{subsec:gpu_optimizations}
During the offline phase, we unroll the~$LDL^\top$ factorization into a program containing the column reach sets, destination indices for $L$, row-update lists, permutation tables, and backsolve tables.
As columns of the factorization may have different amounts of structural nonzeros, the lists are padded and masked.
At runtime, the GPU receives only numerical KKT values, right-hand sides, and the integer tables.
For each batch element, the same symbolic operations are applied to different numerical values and the factorization becomes a function of only the structural nonzeros in~$\bar{\K}$.
We propose three optimizations for
the fixed-pattern program.

\fakepar{Segmented padded scans.}
Naively padding every column to the global maximum work-list length wastes arithmetic when only a few columns attain that maximum.
Let $w_c$ be the number of active process entries in factorization column $c$, see~\eqref{eq:ldl-colum}-\eqref{eq:qdldl_recursion_2}.
A rectangular scan when padding all columns to the maximum costs $W_{\rm global}=(n+m)\cdot\max_c w_c $.
As visualized in Fig.~\ref{fig:segmentation} for three realistic MPC examples, the number of wasted entries and computations can be well over \SI{80}{\percent} of the total when padding uniformly to the maximum.
Instead, we reduce the amount of wasted computations by partitioning consecutive columns into index intervals $\mathcal{I}_\kappa$ and compile each segment with its own local width, giving
\begin{equation}
    W_{\rm seg}=\sum_\kappa |\mathcal{I}_\kappa|\max_{c\in \mathcal{I}_\kappa} w_c .
\label{eq:segmented_work}
\end{equation}
The same construction is used for forward and transpose solves with column nonzero counts of $L$.
Choosing the number of segments (``segment budget'') trades off runtime overhead from masked work for compile time and generated code size.
We compute the optimal segment budget allocation by solving the contiguous partitioning problem induced by \eqref{eq:segmented_work}.
For a budget of $n_{\rm seg}$ segments with $n_{\rm seg}\leq n+m$, we compute offline the ordered nonempty intervals
$\mathcal{I}_\kappa=[a_\kappa,a_{\kappa+1})$ satisfying
$0=a_1<\cdots<a_{\hat n_{\rm seg}+1}=n+m$, such that
\begin{equation}
\min_{\{\mathcal{I}_\kappa\}_{\kappa=1}^{\hat n_{\rm seg}}}
\sum_{\kappa=1}^{\hat n_{\rm seg}}
|\mathcal{I}_\kappa|\max_{c\in\mathcal{I}_\kappa} w_c.
\label{eq:segment_partition_opt}
\end{equation}
We can solve~\eqref{eq:segment_partition_opt} exactly using dynamic programming.

\fakepar{Symbolic-major layout.}
The naive baseline batched layout stores work arrays as $(B,\nnz(\cdot))$ with the batch size $B$.
For a fixed symbolic operation, adjacent GPU threads then access values separated by the potentially very large sparse dimension $\nnz(\cdot)$.
We instead store the main factorization and solve arrays as
$a\in\R^{\nnz(\bar{\K})\times B}$,
$L_x\in\R^{\nnz(L)\times B}$, and
$w,D^{-1},h_{\rm rhs}\in\R^{(n+m)\times B}$.
This makes the whole batch slice contiguous in memory for fixed symbolic operations~\eqref{eq:qdldl_recursion_1}-\eqref{eq:qdldl_recursion_2}.
The symbolic operation remains scalar in the sparse index, but is vectorized
over the batch, which greatly improves throughput at high batch sizes.

\fakepar{Level-scheduled triangular solves.}
\definecolor{parallelkernel}{HTML}{1F77B4}
\definecolor{fusedkernel}{HTML}{8FC4FF}
While factorization is the dominant time in a single sparse solve, each SQP iteration requires multiple ADMM iterations (typically \num{25} to \num{100} in our examples) with full backsolves each.
Thus, triangular solves become a non-negligible part of overall wallclock time.
The fixed sparsity pattern exposes potential parallelism inside the sparse triangular solves, which we exploit.
For each nonzero $L_{ic}$, $i>c$, the forward solve has a dependency from column $c$ to column $i$.
In the offline phase, we precompute dependency levels for columns of $L$ such that the same level has no mutual dependencies. 
Forward substitution then scatters updates from all columns of one level in a single batched operation\rev{, where accumulations from different threads into the same destination are implemented as atomic additions to prevent write conflicts.}
Transpose substitution traverses the same levels in reverse.
This reduces the number of sequential column operations from the number of KKT columns to the number of dependency levels.
In our numerical examples, this results in the reduction of dependency levels between \SI{55}{\percent} and \SI{85}{\percent}, visualized in Fig.~\ref{fig:levelsolve}.
This reduction will only translate to meaningful throughput gains if the batch size does not yet fully occupy the GPU, which we evaluate with batch size sweeps in more detail in the benchmarks in Sec.~\ref{sec:bench}.
Additionally, we mitigate the overhead from many kernel launches by fusing the long tail of levels containing one or two columns (Fig.~\ref{fig:levelsolve}, {\color{fusedkernel}$\blacksquare$}).

\begin{figure}[t]
    \centering
    \includegraphics{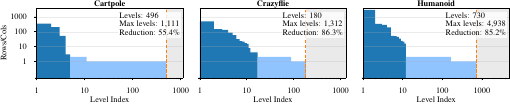}%
    \vspace{-1em}
    \caption{Number of columns per dependency level over dependency level index for the cartpole, quadrotor, and humanoid MPC examples. The number of dependency levels in the backsolve is reduced by \num{55} to over \SI{85}{\percent}, which allows for parallelization ({\color{parallelkernel}$\blacksquare$}) and thus speed ups, when the GPU is not fully utilized. Fusing the long narrow tail ({\color{fusedkernel}$\blacksquare$}) reduces launch overhead.}\label{fig:levelsolve}
    \vspace{-1.5em}
\end{figure}

\begin{table}[b]
    \centering
    \vspace{-1em}
    \caption{Benchmark MPC problem dimensions and solver settings.}
    \vspace{-0.5em}
    \label{tab:benchmarks}
    \resizebox{3.4in}{!}{%
    \setlength{\tabcolsep}{3pt}
    \begin{tabular}{lcccccccccc}
    \toprule
    \textbf{System} & $N$ & $n$ & $m$ & $\operatorname{nnz}(\mathcal{K})$ & \makecell{Fact.\\Instr.} & \makecell{SQP Dat.\\Instr.} & \makecell{SQP\\Iter.} & \makecell{QP\\Iter.} & \makecell{Segm.\\Budg.} & \makecell{Indep.\\Lvls.}\\
    \midrule
    Linear MPC & \num{40} & \num{652} & \num{1144} & \num{9799} & \num{10644} & -- & -- & \num{25} & \num{16} & 310/1796 \\
    Cartpole & \num{100} & \num{505} & \num{606} & \num{4723} & \num{3436} & \num{243} & \num{10} & \num{50} & \num{384} & 496/1111 \\
    Quadrotor & \num{40} & \num{656} & \num{656} & \num{7104} & \num{11687} & \num{617} & \num{1} & \num{25} & \num{256} & 180/1312 \\
    Humanoid & \num{27} & \num{1704} & \num{3234} & \num{70810} & \num{99919} & \num{110075} & \num{1} & \num{25} & \num{96} & 730/4938 \\
    \rev{Juggling} & \num{20} & \num{714} & \num{1848} & \num{70752} & \num{56427} & \num{319173} & \num{2} & \num{40} & \num{256} & 631/2562 \\
    \bottomrule
\end{tabular}
    }
\end{table}

\section{Benchmarks}
\label{sec:bench}
We evaluate the CasADi to JAX translation used for computing SQP problem data in Sec.~\ref{sec:bench_numpysadi}.
We test the unrolled sparse~$LDL^\top$ factorization and backsolve to highlight the effectiveness of memory layout optimization, segmentation, and dependency level parallelization in Sec.~\ref{sec:bench_qdldl}.
We test our ADMM implementation and sensitivity computation on a linear MPC in Sec.~\ref{sec:bench_osqp}.
In Sec.~\ref{sec:bench_sqp}, we provide three realistic, nonlinear MPC benchmarks to measure throughput of \toolboxname in closed-loop simulations.
All benchmark problem dimensions and settings are summarized in Table~\ref{tab:benchmarks}.
Benchmarks are performed on NVIDIA H100 GPUs with 96GB HBM2e, 24 Intel Xeon 8468 Sapphire CPU cores at \SI{2.1}{\giga\hertz}, and in single precision when not explicitly stated otherwise.

\definecolor{paperOursBlueDark}{HTML}{08519C}
\definecolor{paperCuDSSOrange}{HTML}{E69F00}
\definecolor{paperDenseLUMagenta}{HTML}{9E0079}
\definecolor{paperBoxOSQPPurple}{HTML}{7E57C2}
\definecolor{paperMPAXOrangeRed}{HTML}{D55E00}
\definecolor{paperPCGGreen}{HTML}{009E73}
\definecolor{paperCPUGray}{gray}{0.05}
\definecolor{paperMPPIMagenta}{HTML}{CC79A7}
\definecolor{paperCEMBrown}{HTML}{8C564B}

\subsection{Translating CasADi functions to JAX}
\label{sec:bench_numpysadi}
CasADi~\cite{Andersson2019} is a popular software package for sparse differentiation and optimization.
There exists software to automatically generate custom CUDA kernels~\cite{jeon2024cusadi} and JAX functions~\cite{jaxadi2024} from CasADi expressions.
However, using custom CUDA kernels from JAX reduces flexibility and~\cite{jaxadi2024} is limited to small functions, which prohibits the use for large systems like humanoids.
In our humanoid benchmark, computing Jacobians of multibody dynamics constraints can mean over \num{100000} floating point operations per MPC horizon timestep.
To handle such large functions, we implement a custom CasADi to JAX conversion based on string replacement in CasADi SX generated C-code.
This surprisingly provides significant speedups compared to CusADi~\cite{jeon2024cusadi} and scales to very large functions that~\cite{jaxadi2024} cannot handle.

\definecolor{paperNumPySADiOursBlue}{HTML}{1F77B4}
\definecolor{paperJaxADiRed}{HTML}{D62728}
\definecolor{paperCusADiGreen}{HTML}{2CA02C}
\begin{figure}[t]
    \centering
    \includegraphics{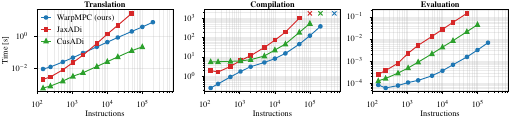}%
    \vspace{-1em}
    \caption{Code export, compilation, and execution time over number of floating point operations in synthetic functions. We compare our CasADi to JAX translation ({\color{paperNumPySADiOursBlue}\ding{108}}) with JaxADi~\cite{jaxadi2024} ({\color{paperJaxADiRed}$\blacksquare$}) and CusADi~\cite{jeon2024cusadi} ({\color{paperCusADiGreen}$\blacktriangle$}). For sweeps across number of instructions we evaluate on batch size \num{10000} and set a compilation time limit to \SI{30}{\minute} ($\times$).
    We significantly outperform the baselines in compilation and evaluation times.
    }
    \label{fig:numpysadi}
    \vspace{-1em}
\end{figure}

As a benchmark, we generate random CasADi functions of different sizes and compare export, compilation, and execution times using our implementation with JaxADi~\cite{jaxadi2024} and CusADi~\cite{jeon2024cusadi}.
We report the results in Fig.~\ref{fig:numpysadi}: 
Our code export with string replacements outperforms both baselines in compilation and evaluation speed.
We argue that the translation disadvantage (Fig.~\ref{fig:numpysadi}, left) is insignificant compared to the overall compilation times.
Importantly, the execution time of the compiled code is an order of magnitude faster than both baselines.
We attribute this speedup to the algorithmic form introduced by the CasADi code generator, which seems to allow the XLA compiler to find better optimizations compared to the nested ``single-line-of-code'' in~\cite{jaxadi2024}.
To achieve this speedup consistently with different JAX versions, it is paramount to disable the multi output fusion optimization\footnote{\texttt{--xla\_disable\_hlo\_passes=multi\_output\_fusion}}.%

\subsection{Optimized Unrolled Sparse Linear Factorization and Solve}
\label{sec:bench_qdldl}
\begin{figure}[t]
    \centering
    \includegraphics[width=\linewidth]{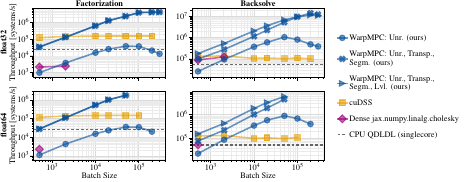}
    \vspace{-2em}
    \caption{Effect of optimizations for a single sparse factorization (left) and backsolve (right) on GPU. We benchmark the throughput over batch size for \toolboxname ({\color{paperOursBlueDark}\ding{108}}, {\color{paperOursBlueDark}\ding{54}}, {\color{paperOursBlueDark}$\blacktriangleright$}) against sparse cuDSS~\cite{nvidia_cudss_preview} ({\color{paperCuDSSOrange}$\blacksquare$}), dense factorization~({\color{paperDenseLUMagenta}\ding{117}}), and CPU factorization ({\color{paperCPUGray}\textbf{- \!-}}), all of which we outperform for batches over \num{2000}.
    }
    \label{fig:results_qdldl}
    \vspace{-1em}
\end{figure}

We benchmark linear system factorization and solve in isolation to highlight the benefits from our proposed optimizations.
In this benchmark, we use a sparse linear system typical for MPC problems and time the throughput in solved systems per second, reporting factorization and backsolve throughput separately in Fig.~\ref{fig:results_qdldl}.
We compare against NVIDIA's sparse cuDSS~\cite{nvidia_cudss_preview} factorization, the dense JAX factorization, and CPU-based factorization implemented by OSQP~\cite{stellato2018osqp}.
Notably, the dense factorization reaches memory limits even for small batch sizes.
The cuDSS throughput reaches peak levels for small batches already.
In comparison, the unrolled, sparse and optimized factorization and backsolve outperform all baselines beyond batch sizes of \num{2000}.
The dependency level scheduling improves throughput for small batch sizes when additional parallelization across underutilized GPU resources is possible.
However, asymptotic performance for very large batches matches the case without level scheduling.
In single precision, throughput noticeably plateaus for batch sizes of over~\num{100000}, while double precision benchmarks only run up to \num{50000} batches due to memory limits.
In summary, the optimized unrolled factorization and backsolve outperform all baselines by up to $27\times$ and $105\times$ in single precision throughput.
An H100 GPU achieves the $167\times$ and $270\times$ single core CPU performance of the optimized QDLDL C implementation.

\subsection{Fixed-Pattern QP Solves and Sensitivities with ADMM}
\label{sec:bench_osqp}
\begin{figure}[t]
    \centering
    \includegraphics[width=\linewidth]{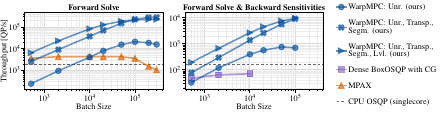}
    \vspace{-2em}
    \caption{Throughput in QPs per second over batch size for a linear MPC without (left) and with (right) sensitivities. We compare \toolboxname's unrolled implementation ({\color{paperOursBlueDark}\ding{108}}) and proposed optimizations ({\color{paperOursBlueDark}\ding{54}}, {\color{paperOursBlueDark}$\blacktriangleright$}) with sparse MPAX~\cite{lu2025mpax} ({\color{paperMPAXOrangeRed}\ding{115}}), dense JAXopt BoxOSQP with CG~\cite{blondel2022efficient} ({\color{paperBoxOSQPPurple}\ding{110}}), and single-threaded CPU OSQP ({\color{paperCPUGray}\textbf{- \!-}}).
    We outperform all baselines for batches over \num{2000}.
    }\label{fig:results_qp}
    \vspace{-1em}
\end{figure}

We test the QP solver implementation and effects of our proposed optimizations in batched linear MPC without the SQP data generation.
We benchmark against a \emph{dense} GPU OSQP implementation BoxOSQP using CG linear solves~\cite{blondel2022efficient}, the sparse MPAX solver~\cite{lu2025mpax} in JAX, and the throughput on a single CPU core with the OSQP C-code implementation~\cite{stellato2018osqp}.
We report the results for single (top) and double (bottom) precision in Fig.~\ref{fig:results_qp} for a naive unrolled factorization and the proposed optimizations with and without scheduling dependency levels in the backsolves.
We report throughput for only solving the MPC problem (left) and additional sensitivities (right) in a backward pass, which requires an additional factorization and backsolves (Sec.~\ref{subsec:qp_sensitivities}).

In this benchmark, we consistently outperform the MPAX baseline (which requires significantly more iterations) and achieve throughputs of \num{60} CPU cores with and \num{150} without sensitivities, which is only possible due to the proposed optimizations.
Peak throughput reaches \num{279000} solves per second without sensitivities and \num{9570} solves per second with sensitivities.
Notably, MPAX computes sensitivities by backpropagating through the computational graph of the forward pass, for which even the linear MPC example is too large.
The \emph{dense} BoxOSQP~\cite{blondel2022efficient} can only handle small batches and does not improve throughput compared to a single CPU core.
The advantage of scheduling backsolve dependency levels diminishes for large batches, in line with Fig.~\ref{fig:results_qdldl}.
\rev{With 25 ADMM iterations and 100 refinement iterations, we achieve normalized sensitivity errors of 7.34\%/36.72\%/53.22\% (median/90th/95th percentile) compared to finite differences, which decrease to 0.135\%/3.914\%/8.523\% when using 100 ADMM iterations with the same 100 refinement iterations.}
\rev{At $10^4/10^5$/max batch size, WarpMPC with all optimizations uses $1.6/16.8/41.6$ GiB (linear) and $5.1/68.9/68.9$ GiB (with sensitivities) of VRAM.}

\subsection{Nonlinear MPC}
\label{sec:bench_sqp}
We benchmark \toolboxname on three nonlinear problems: cartpole swingup and stabilization~\cite{hose2024parameter}, quadrotor setpoint tracking~\cite{barroscarlos2020}, and whole-body humanoid robot locomotion~\cite{khazoom2024tailoring}.
As highlighted in Tab.~\ref{tab:benchmarks}, the three systems have vastly different dimensions (\num{4} to \num{78} states), horizons (\num{27} to \num{100}), and complexities (number of floating point operations in a single SQP data linearization stage $k$ from \num{243} to \num{110075}, number of floating point operations in factorizing $\bar{\K}$ from \num{3436} to \num{99919}).
All systems have nonlinear dynamics and general constraints, e.g., the humanoid has action, self-collision, ground contact, and friction cone constraints, see~\cite{khazoom2024tailoring}.

We compare our implementation and optimizations with sparse preconditioned conjugate gradient (PCG) linear solvers with the Jacobi preconditioner proposed in~\cite{schubiger2020gpu}.
We set maximum PCG iterations to \num{20} and convergence tolerance\footnote{Diverging simulations at $10^{-4}$~PCG~tolerance~from~\cite{adabag2026differentiable}~in~all~examples.} to $10^{-5}$.
We additionally benchmark against MPAX~\cite{lu2025mpax}.
Both PCG and MPAX baselines are implemented as backends in our library, thus using \toolboxname's SQP and line search implementation.
Furthermore, we compare with the concurrent work TurboMPC~\cite{bravopalacios2026turbompc}, using its fused cuDSS backend and otherwise identical settings to our ADMM implementation.
\rev{We also provide MPPI and CEM baselines.}

\begin{figure}[t]
    \includegraphics{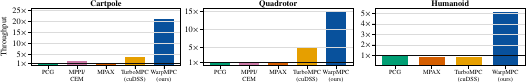}
    \vspace{-2em}
    \caption{Maximum throughput improvement using the optimized \toolboxname~({\color{paperOursBlueDark}\ding{110}}) compared to sparse PCG ({\color{paperPCGGreen}\ding{110}}).
    We also show MPAX ({\color{paperMPAXOrangeRed}\ding{110}}), MPPI/CEM~({\color{paperMPPIMagenta}\ding{110}}), and TurboMPC ({\color{paperCuDSSOrange}\ding{110}}), which we outperform by $3\times$ to $6\times$.
    }
    \label{fig:throughputimprovement}
    \vspace{-1em}
\end{figure}

We find that PCG, MPAX, and TurboMPC (cuDSS) baselines are outperformed beyond batches of \num{10000}.
The comparison with PCG is in line with the results in~\cite{schubiger2020gpu,adabag2024mpcgpu}, where PCG is an effective strategy when accelerating single solves or small batches.
We additionally find improved constraint satisfaction with WarpMPC's sparse factorization compared to PCG.
We attribute this to the PCG solver tolerance. Therefore, improving constraint satisfaction with PCG would come at the expense of throughput.
In all three examples, MPAX has significantly lower closed-loop performance even after \rev{\num{500}} QP iterations compared to ADMM.
The TurboMPC baseline exhibits remarkable performance on small batches, yet fails with out-of-memory errors on large batches in the cartpole and quadrotor examples and does not run for batches beyond \num{512} on the humanoid for the same reason.
\rev{Sampling based methods produce worse solution quality with $1.23\times$ to $2.22\times$ increased cost on cartpole and quadrotor, and fail entirely on the humanoid due to excessive constraint violations.}
The effectiveness of all of \toolboxname's optimizations is noticeable in the nonlinear MPC examples, with the effect of the parallel solve level optimization diminishing for very large batches when the GPU is fully utilized.
In Fig.~\ref{fig:throughputimprovement}, we summarize \toolboxname's effectiveness as improvement over the best PCG throughput observed across all batch sizes.
\toolboxname significantly improves throughput by up to $25\times$.
We outperform TurboMPC by $6\times$ in the cartpole and $3\times$ in the quadrotor example.
\rev{All reported solvers trade off optimality and constraint satisfaction.
We tuned all solvers' settings for similar solution quality and  summarize their primal and dual residuals in Tab.~\ref{tab:primaldual}.}
\rev{At $10^4/10^5$/max batch size, WarpMPC with all optimizations uses $1.7/17.6/36.6$ GiB (cartpole), $3.1/20.6/70.4$ GiB (quadrotor), and $12.0/\text{--}/53.5$ GiB (humanoid) of VRAM.}
\begin{figure}[t!]
    \centering
    \includegraphics[width=\linewidth]{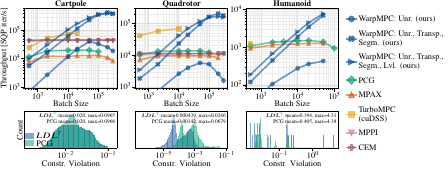}
    \vspace{-2em}
    \caption{Overall throughput in SQP iterations per second over batch size for three nonlinear MPC examples: cartpole pendulum (left), quadrotor (center), and humanoid robot (right). We compare \toolboxname ({\color{paperOursBlueDark}\ding{108}}, {\color{paperOursBlueDark}\ding{54}}, {\color{paperOursBlueDark}$\blacktriangleright$}) with a sparsity exploiting PCG~\cite{schubiger2020gpu} ({\color{paperPCGGreen}\ding{117}}), MPAX~\cite{lu2025mpax} ({\color{paperMPAXOrangeRed}\ding{115}}) for QP solving, TurboMPC~\cite{bravopalacios2026turbompc} ({\color{paperCuDSSOrange}$\blacksquare$}), \rev{and sampling based MPPI and CEM ({\color{paperMPPIMagenta}\ding{116}}, {\color{paperCEMBrown}\ding{58}})}.
    For batch sizes above \num{10000}, \toolboxname consistently improves throughput compared to all baselines.
    }\label{fig:nonlinearsystems}
    \vspace{-1em}
\end{figure}

\begin{table}[b]
\centering
\vspace{-1em}
\caption{\rev{Primal and Dual Residuals (Median and 99th Quantile)}}
\vspace{-0.5em}
\label{tab:primaldual}
\resizebox{3.4in}{!}{%
\setlength{\tabcolsep}{3pt}
\begin{tabular}{@{}l *{12}{c}@{}}
\toprule
& \multicolumn{4}{c}{\textbf{Cartpole}}
& \multicolumn{4}{c}{\textbf{Quadrotor}}
& \multicolumn{4}{c}{\textbf{Humanoid}} \\
\cmidrule(lr){2-5}
\cmidrule(lr){6-9}
\cmidrule(lr){10-13}

\textbf{Baseline}
& \multicolumn{2}{c}{Primal}
& \multicolumn{2}{c}{Dual}
& \multicolumn{2}{c}{Primal}
& \multicolumn{2}{c}{Dual}
& \multicolumn{2}{c}{Primal}
& \multicolumn{2}{c}{Dual} \\

& Median & 99th
& Median & 99th
& Median & 99th
& Median & 99th
& Median & 99th
& Median & 99th \\
\midrule

WarpMPC
& 0.159 & \underline{0.425}
& \underline{0.0247} & \underline{0.0486}
& 0.00023 & \underline{0.015}
& \textbf{0.000001} & \textbf{0.00006}
& \textbf{0.060} & \textbf{0.34}
& \textbf{0.0001} & \textbf{0.002} \\

PCG
& 0.159 & 0.426
& 0.0248 & 0.0487
& 0.0012 & 0.043
& 0.00014 & \underline{0.0062}
& \underline{0.14} & 0.89
& 0.40 & 1.7 \\

MPAX
& \underline{0.0447} & \textbf{0.354}
& 0.0548 & 0.637
& \underline{0.000097} & 0.017
& \underline{0.000058} & 0.014
& 0.56 & 2.8
& 0.11 & 1.1 \\

TurboMPC
& \textbf{0.0042} & 0.653
& \textbf{0.0035} & \textbf{0.047}
& \textbf{0.000001} & \textbf{0.0002}
& 0.32 & 1.3
& 0.20 & \underline{0.43}
& \underline{0.0013} & \underline{0.0033} \\

\bottomrule
\end{tabular}
}%
\end{table}

\begin{figure}[t]
    \centering
    \includegraphics[height=0.9in, trim=550 50 550 50, clip]{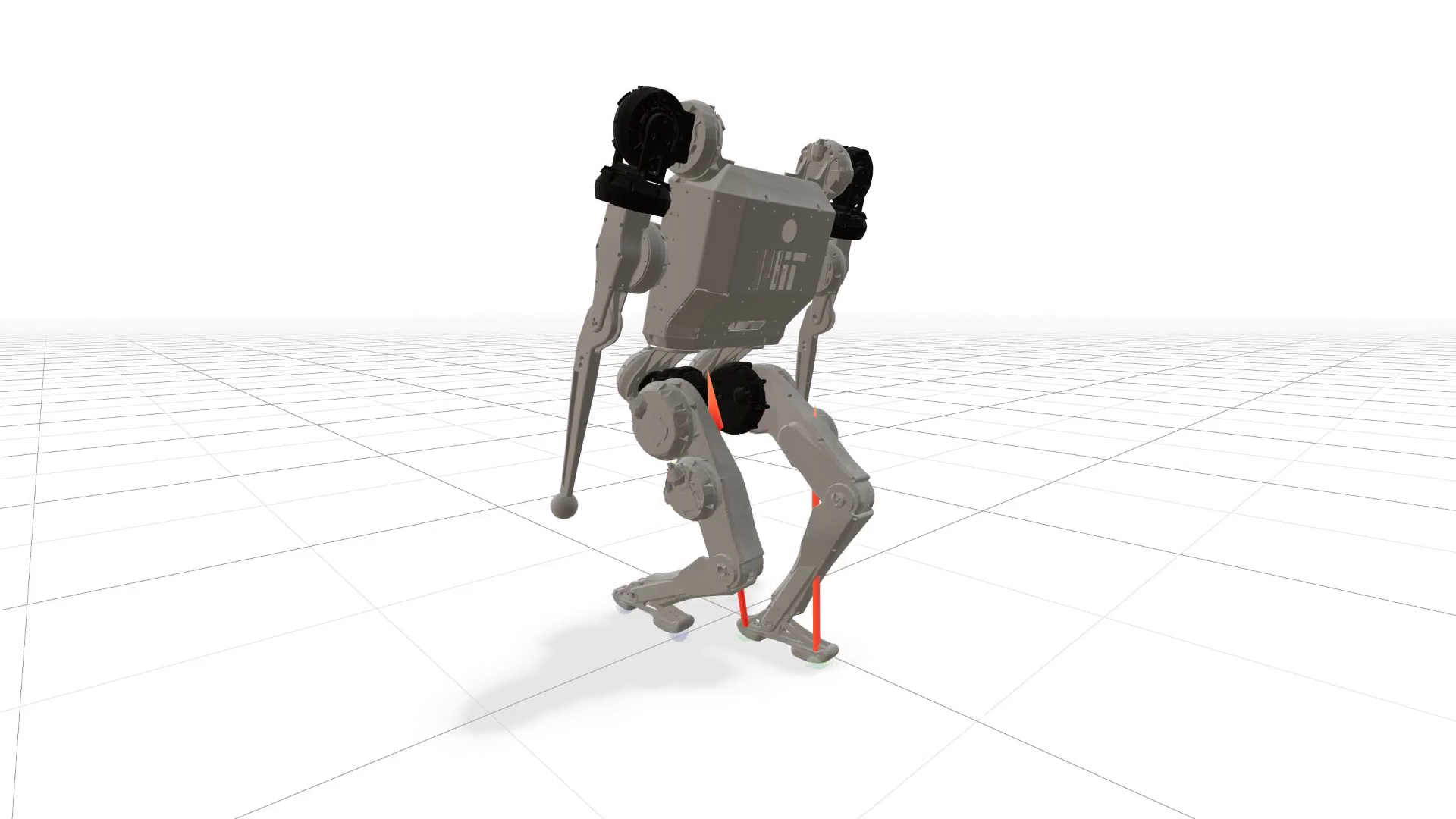}%
    \includegraphics[height=0.9in, trim=550 50 550 50, clip]{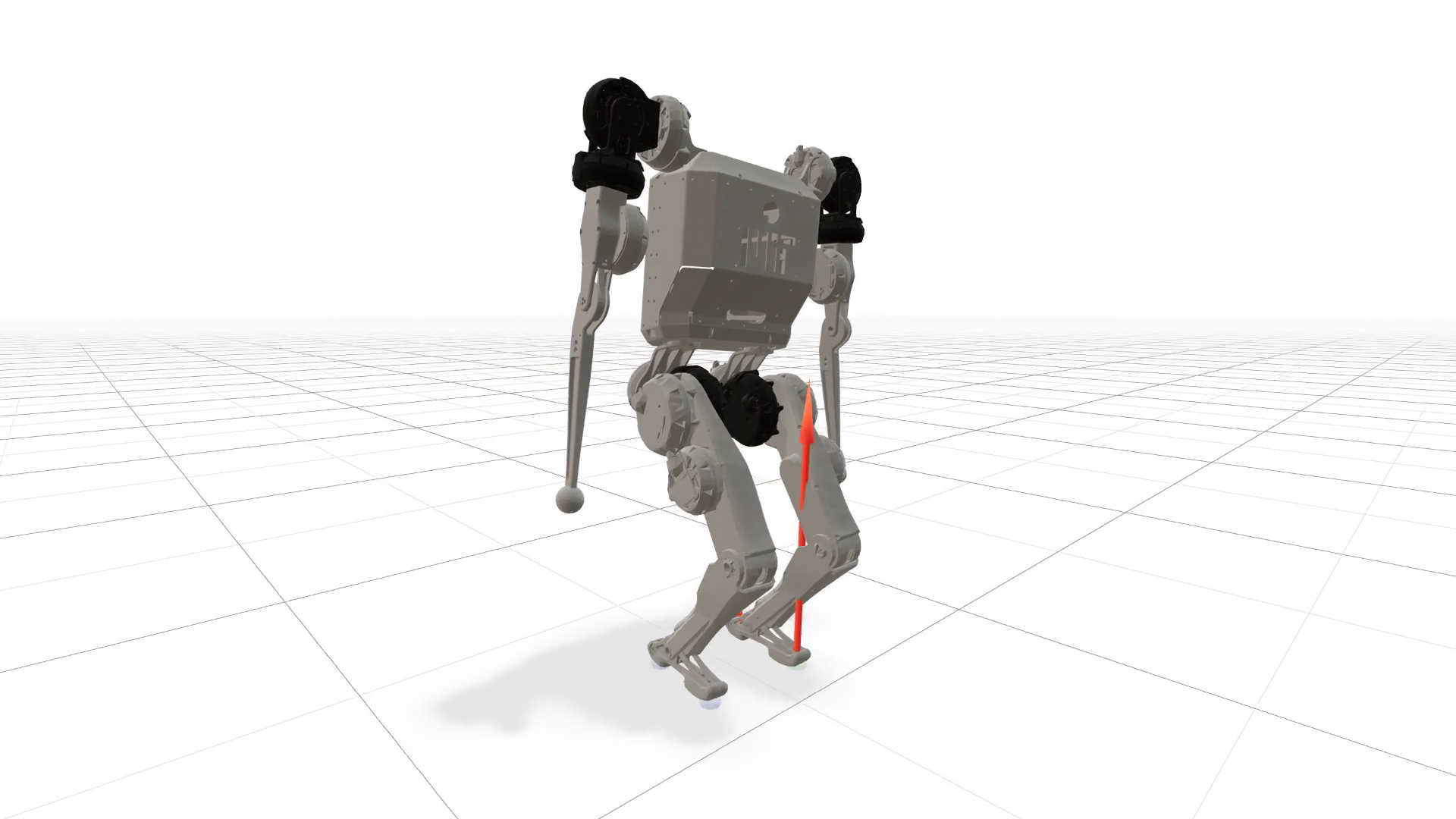}%
    \includegraphics[height=0.9in, trim=0 0 0 150,clip]{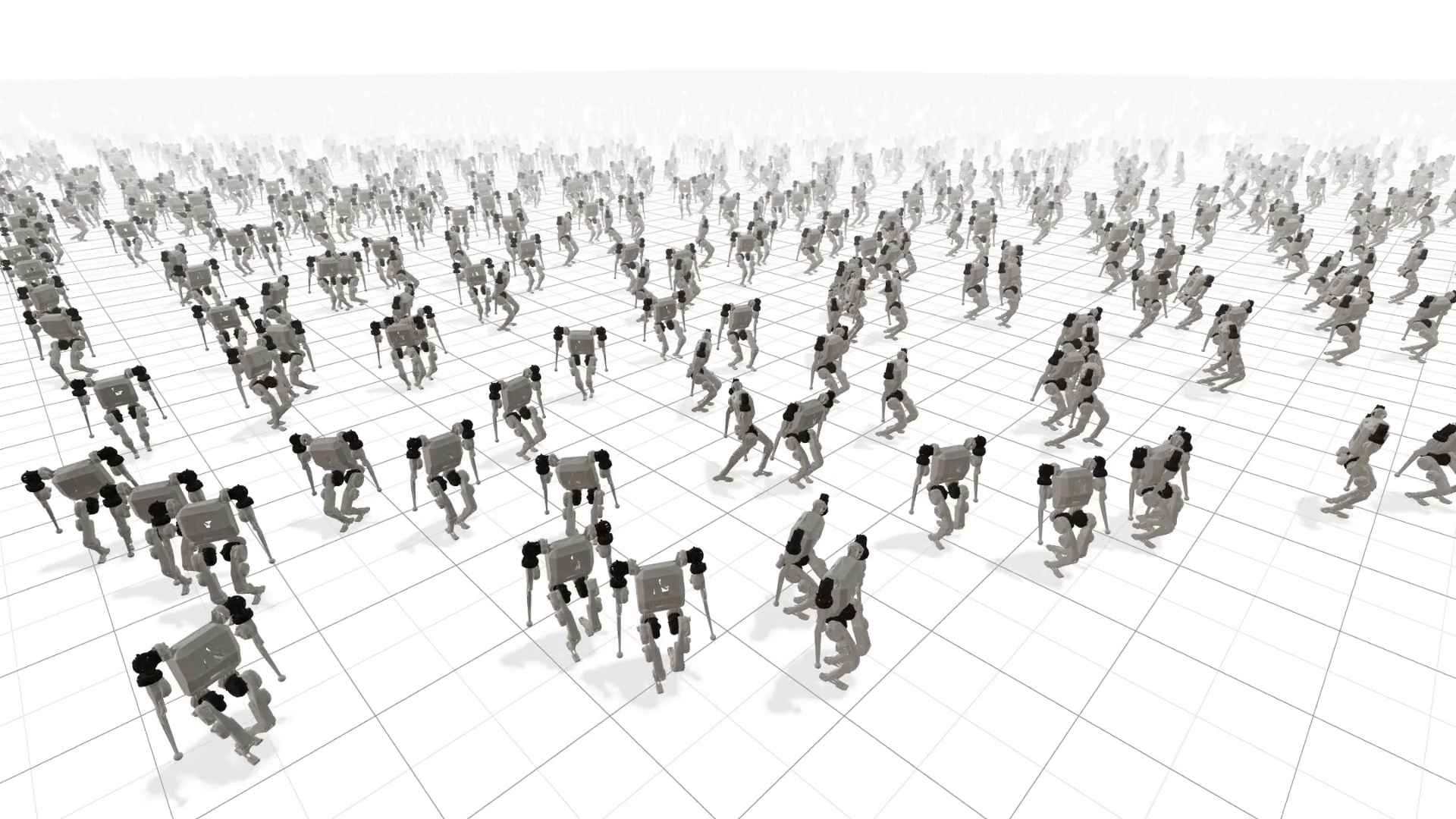}
    \vspace{-1em}
    \caption{Humanoid whole-body MPC with forward (left) and sideways (center) walking, and \num{10000} parallel environments (right). The MPC formulation from~\cite{khazoom2024tailoring} implemented with \toolboxname on GPU stabilizes a walking gait in simulation with hard impulse-based contacts~\cite{stewart2000implicit}.}
    \label{fig:humanoids}
    \vspace{-1em}
\end{figure}

In comparison to~\cite{jeon2025residual} reporting humanoid MPC throughputs of \num{2000} SQP RTI/s, we achieve almost \num{8000} SQP RTI/s on a more complex MPC matching~\cite{khazoom2024tailoring}, see rendering in Fig.~\ref{fig:humanoids}.
The KKT system in~\cite{jeon2025residual} is smaller due to the horizon length of only \num{12} compared to our $N\!=\!27$.
We highlight that \toolboxname's throughput is only possible due to larger batches.

\begin{figure}
    \centering
    \includegraphics{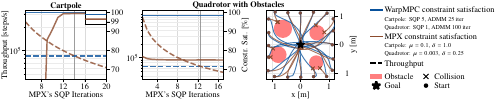}
    \vspace{-0.5em}
    \caption{Constraint satisfaction and throughput in simulation steps per second over SQP iterations of MPX and \toolboxname for cartpole (left) and quadrotor setpoint tracking with additional obstacles (center), for which trajectories are shown (right) of MPX with 20 and \toolboxname with 1 SQP iteration.
    }
    \label{fig:mpxcomparison}
    \vspace{-1em}
\end{figure}

The iLQR method MPX~\cite{amatucci2025primal} cannot directly handle constraints, but requires encoding these in the cost~\cite[Eqn. (21)]{amatucci2025primal} with tuned penalties.
Nonetheless, MPX shows exceptional throughput due to parallelization in the horizon.
We report MPX's percentage of constraint satisfying simulations over the number of SQP iterations for cartpole and a modified quadrotor with cylindrical obstacles in Fig.~\ref{fig:mpxcomparison}.
For \toolboxname, almost~\SI{100}{\percent} constraint satisfaction is possible with very few SQP iterations.
At throughput parity, MPX shows significantly worse constraint satisfaction on the quadrotor with obstacles.
For the cartpole example, MPX plateaus at high constraint satisfaction just below that of 5 SQP iterations with~\toolboxname.

\section{Illustrative Example: Neural Network MPC}

\label{sec:ampc}
The limited computing resources and fast system dynamics make deployment of MPC difficult, e.g., on nano quadrotors, such as the Crazyflie.
A way to speed up computations is to synthesize a large dataset of states and optimal actions from an MPC in offline simulations, train a neural network in a supervised learning setup to explicitly approximate the optimal actions, and deploy this neural network on the embedded target. %
Prior work was limited to low-dimensional systems~\cite{hertneck2018learning,lucia2020deep} or required large computing clusters to naively parallelize with existing solvers on CPU~\cite{hose2024parameter,julbe2025diffusion}.
With \toolboxname, we can now synthesize large datasets and train neural networks on a single GPU within minutes.
\rev{We provide two such examples: quadrotor setpoint control directly controlling thrusts in the stabilizer loop of a Crazyflie and pingpong juggling on a Franka arm with a contact implicit MPC controlling joint torques.
The Crazyflie policy has 3 hidden layers with 32 neurons per layer and leaky ReLU activations and is trained on 20 million state-action pairs.
We synthesize the dataset in \SI{2}{\minute} and train the policy in approximately \SI{1.4}{\minute} wallclock time.
Additional overhead is \SI{0.8}{\minute} for a cold run and \SI{0.13}{\minute} with compilation caching.
The Franka robot pingpong juggling policy has 3 layers with 256 neurons each and is trained on 15 million state-action pairs. Dataset synthesis takes \SI{44}{\minute} (more complicated MPC problem, see Tab.~\ref{tab:benchmarks}) and training takes \SI{1.8}{\minute}.
}

\begin{figure}[bt]
    \vspace{-1em}
    \centering
        \begin{tikzpicture}

        \node[
            rotate=90,
            font=\scriptsize\bfseries,
            anchor=center,
            inner sep=0pt
        ] (cftitle) at (0,0) {Crazyflie};

        \node[
            anchor=west,
            inner sep=0pt
        ] (cfphoto) at ([xshift=0.1in]cftitle) {%
            \includegraphics[
                height=0.8in,
                trim=450 180 180 350,
                clip
            ]{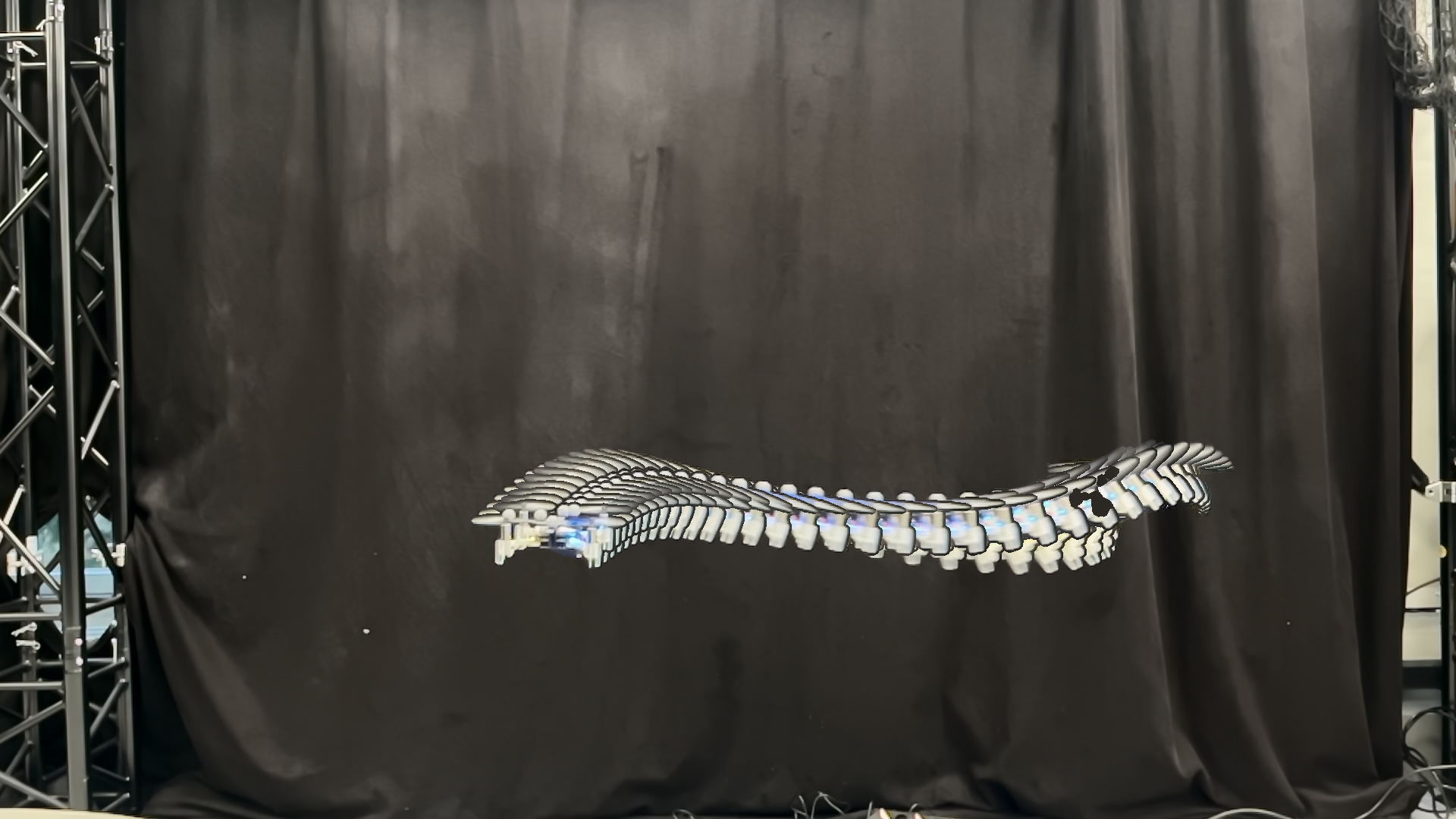}%
        };

        \node[
            anchor=west,
            inner sep=0pt
        ] (cfplot) at ([xshift=0.08in]cfphoto.east) {%
            \includegraphics{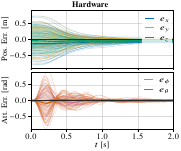}%
        };

        \node[
            anchor=north west,
            inner sep=0pt
        ] (fjleft) at ([yshift=-0.05in]cfphoto.south west) {%
            \includegraphics[height=0.8in, clip, trim=320 0 346 0]{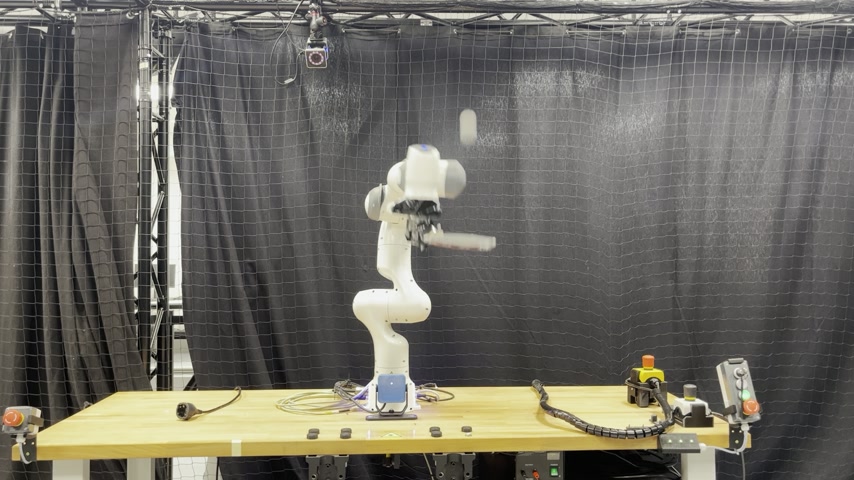}%
            \includegraphics[height=0.8in, clip, trim=320 0 346 0]{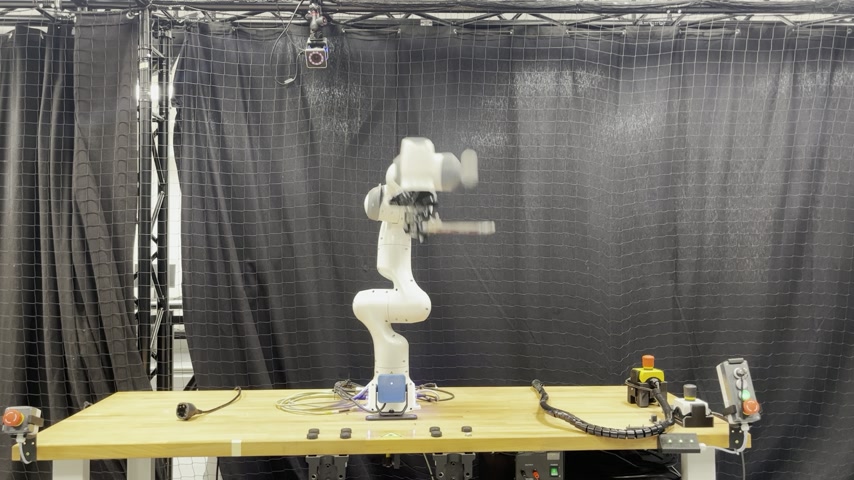}%
            \includegraphics[height=0.8in, clip, trim=320 0 346 0]{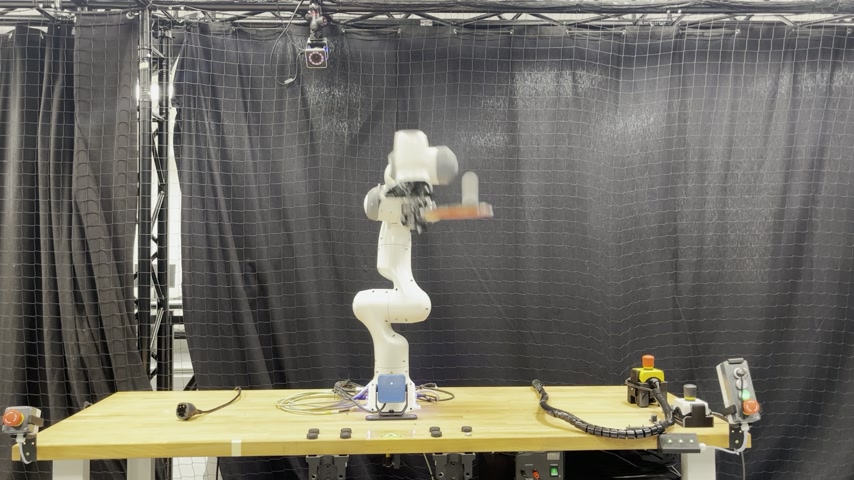}%
            \includegraphics[height=0.8in, clip, trim=320 0 346 0]{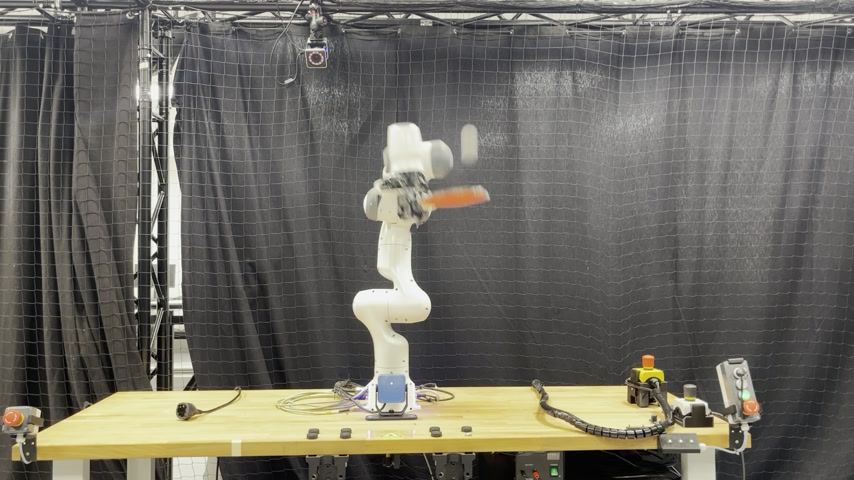}%
            \includegraphics[height=0.8in, clip, trim=320 0 346 0]{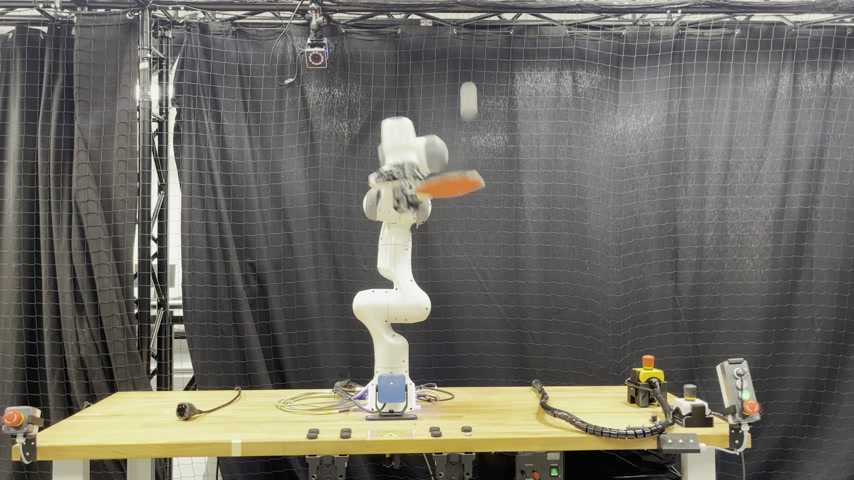}%
            \includegraphics[height=0.8in, clip, trim=320 0 347 0]{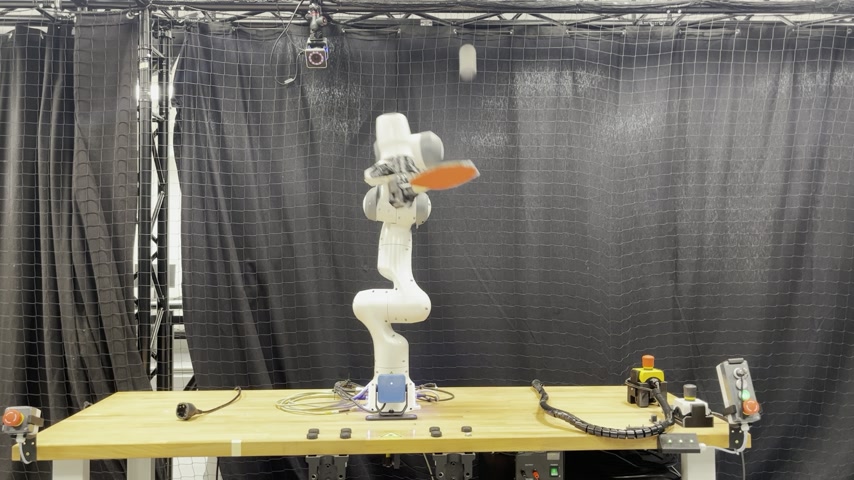}%
        };
        \node[
            anchor=west,
            inner sep=0pt
        ] (fjright) at ([xshift=0.08in]fjleft.east) {%
            \includegraphics{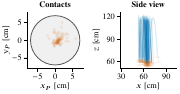}%
        };

        \node[
            rotate=90,
            font=\scriptsize\bfseries,
            anchor=center,
            inner sep=0pt,
            align=center
        ] (fjtitle) at ([xshift=-0.10in]fjleft.west |- fjleft.center) {\rev{Franka Juggling}};

    \end{tikzpicture}
    \caption{Hardware experiments controlling a Crazyflie quadrotor \rev{(top) and Franka pingpong juggling (bottom)} with a neural network approximation of MPC (left: frames overlaid from experiment video, right: state error evolution for tracking randomized setpoints, \rev{paddle hit distribution and ball trajectories}).}
\end{figure}

\section{Conclusion \& Outlook}
We presented \toolboxname, a JAX toolbox with Warp backend for solving large batches of constrained nonlinear MPC problems on GPU via SQP, ADMM, and problem-specific unrolled sparse~$LDL^\top$ factorization.
We propose throughput maximizing optimizations: automatic JAX function generation from CasADi expressions, a memory layout optimization, an optimal segmentation for the factorization, and parallelizing dependency levels in the backsolve.
We achieve throughputs of \num{8000} to \num{250000} SQP iterations per second on large batches of realistic MPC benchmarks, outperforming baselines by $3\times$ to $25\times$.
We illustrate \toolboxname's usefulness by synthesizing a neural network approximation of an MPC for embedded hardware deployment onboard a nano quadrotor and contact-implicit pingpong juggling.
Many common features of CPU based MPC toolboxes are left for future work, e.g., Hessian regularization, and GPU native interfaces for neural network models in the MPC constraints.
Additional optimizations like reordering to minimize padding rather than infill could further improve overall throughput.

\ifarxiv
\section*{Acknowledgments}
We thank J. Terrazzano for feedback on the manuscript.
\fi

\bibliographystyle{IEEEtran}
\bibliography{references}

\end{document}